\documentclass[11pt,letterpaper]{article}
\usepackage[margin=1in]{geometry}
\usepackage{microtype}
\usepackage{graphicx}
\graphicspath{{figures/}{./}}
\usepackage{booktabs,adjustbox,multirow,longtable}
\usepackage[section]{placeins}
\usepackage{xcolor}
\usepackage[round,authoryear]{natbib}
\usepackage{amsmath,amssymb,mathtools,amsthm}
\usepackage{algorithm,algorithmic}
\usepackage[font=small,labelfont=bf]{caption}
\usepackage[hidelinks]{hyperref}
\usepackage[capitalize,noabbrev]{cleveref}

\definecolor{qdeq}{HTML}{D55E00}
\newcommand{\qdeq}{Q-DEQ}
\newcommand{\best}[1]{\textbf{#1}}
\hypersetup{
 pdftitle={Q-DEQ: Discrete Update Decisions for Deep Equilibrium Models on Coherent Ising Machines},
 pdfauthor={Ruotong Yang, Hongdong Zhu, Qi Gao, Yin Ma, Hai Wei, Kai Wen},
 pdfsubject={Deep equilibrium models and discrete optimization},
 pdfkeywords={deep equilibrium models, QUBO, coherent Ising machines, time series forecasting}
}
\title{Q-DEQ: Discrete Update Decisions for Deep Equilibrium Models on Coherent Ising Machines}
\author{Ruotong Yang\thanks{Corresponding author: \href{mailto:renata.rt.1027@gmail.com}{\texttt{renata.rt.1027@gmail.com}}.}, Hongdong Zhu, Qi Gao\\
Yin Ma, Hai Wei, Kai Wen\\[0.4em]
\small Beijing QBoson Quantum Technology Co., Ltd., Beijing, China}
\date{November 23, 2026}

\makeatletter
\renewcommand{\maketitle}{%
 \begin{center}
 {\fontsize{18}{21}\selectfont\bfseries\@title\par}
 \vspace{8pt}
 {\normalsize\@author\par}
 \vspace{5pt}
 {\small\@date\par}
 \end{center}
 \@thanks
 \setcounter{footnote}{0}
}
\makeatother
\renewenvironment{abstract}{\par\small\noindent\textbf{Abstract.}\ }{\par\vspace{8pt}}

\begin{document}
\maketitle
\begin{abstract}
Deep equilibrium models (DEQs) obtain implicit depth from a fixed-point solve, but their continuous update rules provide no direct interface to Ising processors. We introduce \qdeq, which expresses each update as a compact quadratic unconstrained binary optimization (QUBO) problem. History-based directions and finite-difference probes define a local residual objective, and binary codes select the step coefficients. The QUBO size depends on the direction count and coefficient resolution rather than the hidden-state dimension: our default uses 80 binary variables. On five iTransformer forecasting benchmarks over three seeds, \qdeq{} achieves MSE within $-0.53\%$ to $+0.78\%$ of Broyden DEQs; a continuous-coefficient control yields nearly identical forecast error. Physical coherent Ising machine (CIM) execution preserves forecast MSE within 0.35\% of the SA references. Adaptive subspaces reduce network evaluations by 37.7--49.9\% with relative MSE changes between $-0.104\%$ and $+0.011\%$. A residual-descent bound separates model, coefficient-resolution and backend errors. Together, these results establish a discrete interface for heterogeneous DEQ execution, complemented by weight tying and a terminal W8A8 representation.

\end{abstract}

\begin{figure}[!htbp]
\centering
\includegraphics[width=\textwidth,height=0.68\textheight,keepaspectratio]{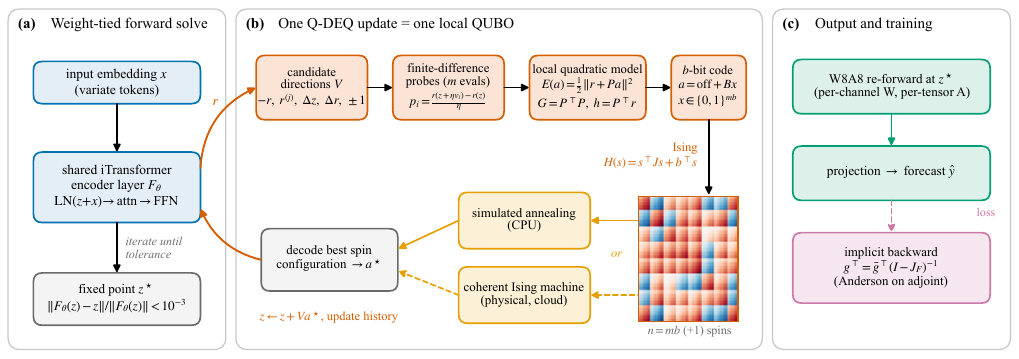}

\caption{\textbf{\qdeq{} overview.} (a) A weight-tied iTransformer layer is iterated to a fixed point. (b) Each update builds $m$ candidate directions from the residual and the history, estimates their effect on the residual with $m$ finite-difference probes, and minimizes the resulting quadratic model over $b$-bit coefficient codes. This is a QUBO with $mb$ variables, solved by SA or by a physical CIM. (c) The forecast is produced by a W8A8 fake-quantized re-forward at the fixed point, and gradients use implicit differentiation.}
\label{fig:overview}

\end{figure}

\section{Introduction}
\label{sec:intro}

Deep equilibrium models (DEQs) represent depth through repeated computation with a single shared layer, finding $z^\star=F_\theta(z^\star,x)$ and training by implicit differentiation \citep{bai2019deq}. This decouples model depth from parameter count, an attractive property for forecasting models whose storage grows with explicit depth \citep{wang2025lightweight,ling2025automating}. The forward solver is central to this representation: Anderson acceleration \citep{anderson1965iterative} and Broyden's method \citep{broyden1965class} use the residual history to choose continuous updates. This raises a hardware-design question: how can the update decision itself be expressed in a form executable by a different class of processor?

Coherent Ising machines (CIMs) select low-energy configurations of coupled spins through the physical evolution of optical parametric oscillators \citep{mcmahon2016fully,inagaki2016coherent,wei2026versatile}. Their native interface is a discrete quadratic objective. A DEQ state, in contrast, is high dimensional and continuous. A useful interface therefore needs to extract a compact discrete decision while retaining the network's continuous representation and the residual information that guides the solve.

\qdeq{} provides this interface by discretizing the coefficients of a low-dimensional update subspace (\cref{fig:overview}). History-based directions identify candidate displacements; finite differences estimate their effect on the nonlinear residual. Minimizing the resulting quadratic model over binary-encoded coefficients produces a QUBO with $mb$ variables for $m$ directions and $b$ bits per coefficient. Its size is independent of the hidden-state dimension, so the same 80-variable decision format serves all five forecasting datasets. A host processor evaluates the network and constructs the objective, while SA or a CIM returns the coefficient code. Low-bit encoding applies to the update decision; an FP32 solve and a terminal W8A8 re-forward complete the mixed-precision computation.

This formulation makes coefficient resolution, backend solution quality and local-model fidelity explicit design variables. It also admits adaptive direction selection and learned initialization within the same discrete interface. We evaluate these choices with continuous-coefficient controls, three-seed forecasting comparisons and physical CIM execution. Our contributions are:
\begin{itemize}
\setlength\itemsep{0.1em}
\item \textbf{A compact discrete interface for equilibrium updates.} We derive a QUBO from history-based directions and a finite-difference residual model, enabling Ising execution inside a DEQ forward solve with implicit training.
\item \textbf{Accuracy-preserving coefficient discretization.} Across five forecasting benchmarks, \qdeq-INT8 is within $-0.53\%$ to $+0.78\%$ MSE of Broyden DEQs. A continuous-coefficient control produces nearly identical forecast error. Weight tying and the W8A8 representation provide complementary logical storage reductions of $4.4$--$12.8\times$ relative to explicit FP32 models.
\item \textbf{Physical execution and adaptable computation.} CIM-driven inference preserves forecast MSE within 0.35\% of the SA references on five datasets. Adaptive subspaces cut network evaluations by 37.7--49.9\% with relative MSE changes between $-0.104\%$ and $+0.011\%$.
\item \textbf{An interpretable error decomposition.} We bound the effects of finite differences, coefficient resolution and backend optimization on residual descent, giving sufficient descent conditions and practical guidance for choosing the discrete interface.
\end{itemize}

\section{Related Work}
\label{sec:related}

\textbf{DEQ solvers.} DEQs \citep{bai2019deq} have been extended to multiscale vision models \citep{bai2020multiscale} and optical flow \citep{bai2022deqflow}. TorchDEQ \citep{geng2023torchdeq} shows that the forward solver, be it fixed-point iteration, Anderson or Broyden, together with the backward solver and stabilization, strongly affects accuracy and efficiency. We compare against both Anderson and limited-memory Broyden with a common implicit backward pass.

\textbf{DEQs with guarantees.} Monotone operator equilibrium networks \citep{winston2020monotone} and positive concave DEQs \citep{gabor2024pcdeq} restrict the layer so that the fixed point exists, is unique, and is reached by suitable iterations. \cref{sec:discussion} discusses how \qdeq{} can be combined with these structures.

\textbf{Ising machines and combinatorial solvers in learning.} Coherent Ising machines solve Ising problems through the dynamics of degenerate optical parametric oscillators \citep{mcmahon2016fully,inagaki2016coherent,wei2026versatile}. QUBO modelling tools such as PyQUBO \citep{zaman2021pyqubo} map applications onto such hardware, and black-box differentiation \citep{vlastelica2020blackbox,rolinek2020bbgm} embeds combinatorial solvers as network layers. In \qdeq{} the combinatorial problem is an inner step of a fixed-point solver rather than a task output.

\textbf{Quantizing weight-tied models.} Recent work studies low-bit quantization of recurrently reused modules \citep{ingolfsson2026quantizing,fang2026loopq,jim2026survives}. These methods quantize weights and activations. \qdeq{} discretizes update coefficients and separately uses quantization-aware training for the terminal W8A8 re-forward \citep{jacob2018quantization}.

\textbf{Learned accelerators and stabilization.} Consistency DEQs \citep{lin2026consistency} distill solver trajectories into a time-conditioned backbone using global and local consistency. Learned update predictors amortize solver decisions through training; Q-DEQ constructs a residual objective at the current state and selects an encoded update without an additional learned coefficient predictor. This explicit objective provides the interface to an Ising backend. The two mechanisms compose: our forecasting adaptation of C-DEQ supplies an initialization for discrete refinement (\cref{sec:computation}). Jacobian regularization \citep{bai2021stabilizing} improves map conditioning, while history-based directions and adaptive subsets reuse multisecant information with reduced redundancy.

\textbf{Residual-aware acceleration.} Physics-aware Anderson acceleration for hybrid PDE solvers \citep{wu2026reliable} selects mixing coefficients using the physical equation residual rather than the learned update magnitude. This shares the principle of choosing coefficients through a residual objective. Q-DEQ models the nonlinear DEQ residual $F_\theta(z)-z$ along candidate directions and encodes the coefficients for discrete execution.

\section{Method}
\label{sec:method}

\textbf{Fixed-point model.} $F_\theta$ is one iTransformer encoder layer acting on variate tokens; the embedded input $x$ is re-injected as $\mathrm{LN}(z+x)$ and the iteration starts at $z^{(0)}=x$. A solve stops when $\rho(z)=\|F_\theta(z,x)-z\|/(\|F_\theta(z,x)\|+\delta)$ falls below $\epsilon=10^{-3}$ or its step cap is reached; the Ising implementation uses $\delta=10^{-9}$. Residuals and denominators are evaluated at the same iterate.

\textbf{Candidate directions.} With residual $r=F_\theta(z)-z$, \qdeq{} forms $m$ unit-norm directions $V=[v_1,\dots,v_m]$: $v_1=-r/\|r\|$, then the most recent residuals, then recent pairs $(\Delta z,\Delta r)$ from a history of four steps, and Rademacher vectors for any remaining slots.

\textbf{Local model.} The change of the residual along $v_i$ is estimated by a forward difference with step $\eta$,
\begin{equation}
p_i=\big(r(z+\eta v_i)-r(z)\big)/\eta ,
\end{equation}
costing one evaluation of $F_\theta$ per direction; dropout is disabled during the solve. For coefficients $a\in\mathbb R^m$ the residual after $z\leftarrow z+Va$ is modelled as $r+Pa$, so
\begin{equation}
E(a)=\tfrac12\|r+Pa\|^2-\tfrac12\|r\|^2=\tfrac12 a^\top G a+h^\top a,
\end{equation}
with $G=P^\top P$ and $h=P^\top r$. $E(a)<0$ predicts a decrease of the residual.

\textbf{Binary encoding.} Each $a_i$ lives on a uniform $2^b$-level grid over $[-Rs,Rs]$ with $s=\|r\|/\sqrt m$, that is, $a=\mathrm{off}+Bx$ with $x\in\{0,1\}^{mb}$. The continuous coefficient box contains the Picard displacement through $a_1=-\|r\|$ when $R\geq\sqrt m$; a finite uniform grid approximates this coefficient. The explicit safeguard fallback is evaluated directly in state space. Substituting gives the QUBO
\begin{equation}
\min_{x\in\{0,1\}^{n}}\; x^\top Qx+q^\top x ,
\end{equation}
with $Q=\tfrac12B^\top GB$, $q=B^\top(G\,\mathrm{off}+h)$ and $n=mb$. It maps to an Ising Hamiltonian $s^\top Js+b^\top s$ over $n+1$ spins, where one auxiliary spin carries the linear terms. The returned configurations are decoded and the one with the lowest $E$ is applied. We use $m=8$ and $b=10$, hence $n=80$, throughout; $R$ and $\eta$ are set per dataset as listed in \cref{app:hparams}.

\begin{algorithm}[t]
\caption{Q-DEQ with optional damping and safeguard}
\label{alg:qdeq}
\begin{algorithmic}[1]
\STATE $z\gets x$, $H\gets\emptyset$; damping $\lambda\geq0$; safeguard $\sigma\in[0,1]$ (default $0$)
\FOR{$k=1,\ldots,K$}
\STATE $u\gets F_\theta(z,x)$; $r\gets u-z$; $\rho\gets\|r\|/(\|u\|+10^{-9})$
\IF{$\rho<\epsilon$} \STATE \textbf{break} \ENDIF
\STATE $V\gets\textsc{Directions}(r,H)$; choose probe step $\eta$
\STATE $p_i\gets[F_\theta(z+\eta v_i,x)-(z+\eta v_i)-r]/\eta$
\STATE $G\gets P^\top P$; $h\gets P^\top r$; $\widetilde G\gets G+\lambda\operatorname{tr}(G)I/m$
\STATE choose $R$; $(Q,q)\gets\textsc{Encode}(\widetilde G,h,R,b)$
\STATE $\xi^\star\gets\textsc{SA}(Q,q)$ or $\textsc{CIM}(Q,q)$
\STATE $a\gets\mathrm{off}+B\xi^\star$; $z_c\gets z+Va$; $r_c\gets F_\theta(z_c,x)-z_c$
\IF{$\sigma>0$ and $\|r_c\|\geq\sigma\|r\|$}
\STATE $z_c\gets u$; $r_c\gets F_\theta(z_c,x)-z_c$ \hfill Picard fallback
\ENDIF
\STATE update $H$ with $(r,z_c-z,r_c-r)$; $z\gets z_c$
\ENDFOR
\STATE \textbf{return} a W8A8 re-forward at $z$
\end{algorithmic}
\end{algorithm}

\textbf{Safeguard and returned state.} The main SA results use $\sigma=0$. The safeguard compares absolute residual norms; $\sigma=1$ requires a decrease, and $0<\sigma<1$ requires a fixed contraction factor. Rejection triggers Picard fallback. The Traffic hardware profile also returns the best accepted outer iterate by relative residual. \Cref{app:policies} consolidates the policies and their guarantee conditions.

\textbf{Parameter selection and online control.} The coefficient range controls step length, while $\eta$ controls the locality and numerical resolution of the probes. Fixed values are selected by held-out validation performance; \cref{tab:hparams} consolidates all dataset settings. Weather adapts the box to the current continuous local-model solution,
\[
R_k=\operatorname{clip}_{[0.5,40]}\bigl(1.6\|a_{\rm cont}\|_\infty/s\bigr),
\]
which covers its largest coefficient with a validation-selected margin. ETTh2 uses
\[\eta_k=\max(0.45R_ks,10^{-4}),\]
sampling at a scale tied to the permitted update. The other datasets use fixed probe steps. Both rules depend on the current input and residual model. Data normalization is fitted on the training split, and early stopping uses validation loss. A separate validation-only sensitivity study is reported in \cref{app:round3}.

\textbf{Accuracy of a discrete local step.} Let $r=F(z)-z$, with unit-norm columns in $V$, and suppose the residual Jacobian is $L$-Lipschitz along the probe and update segments. For forward differences with step $\eta$, define
\begin{equation}
D(a)=\tfrac{L\eta}{2}\|a\|_1+\tfrac L2\|Va\|^2.
\label{eq:model-defect}
\end{equation}
Then $\|r(z+Va)\|\leq\|r+Pa\|+D(a)$. If $a_\Box$ minimizes the continuous model on the coefficient box and $\widehat a$ has objective gap at most $\varepsilon_Q$ from the best grid point, then
\begin{equation}
\begin{aligned}
\|r(z+V\widehat a)\|\leq{}&\|r+Pa_\Box\|+\frac{R\|P\|_2\|r\|}{2^b-1}\\
&+\sqrt{2\varepsilon_Q}+D(\widehat a).
\end{aligned}
\label{eq:discrete-bound}
\end{equation}
The terms expose four design levers: subspace fit, coefficient resolution, backend quality and local-model fidelity. Increasing $b$ reduces the grid term; $R$ balances step reach against nonlinear-model error. A right-hand side below $\sigma\|r\|$, $\sigma<1$, guarantees residual contraction even for a noncontractive $F$. \Cref{app:model-bound} gives the proof, bounded-noise extension and conditions for repeated-step convergence; \cref{app:trained-terms} measures directional model defects on frozen trained maps.

\textbf{Quantized re-forward and training.} The solve runs at full precision. At the returned state the shared layer is evaluated once more with W8A8 fake quantization, using symmetric per-output-channel weights and per-tensor dynamic activations clipped to $[-127,127]$, with straight-through gradients \citep{bengio2013estimating}. Gradients use implicit differentiation, solving the adjoint equation with Anderson acceleration. DEQ models share an Anderson-DEQ warm-up initialization. All DEQ variants use the same implicit backward procedure; dataset-specific forward and backward iteration caps are listed in \cref{tab:hparams}.

\section{Experimental Setup}
\label{sec:setup}

We use the Weather, ETTh1, ETTh2, ECL and Traffic benchmarks \citep{zhou2021informer,wu2021autoformer}, which have between 7 and 862 variates, with input length 96 and horizon 96. Explicit baselines follow the reference iTransformer experimental configurations \citep{liu2024itransformer}, with two to four layers. We retain these configurations as fixed forecasting reference points and use Anderson/Broyden DEQs to assess the discrete solver within a shared-layer architecture. All DEQ variants share one encoder layer of the same width, use a maximum of 10 epochs with patience 3, and select checkpoints by validation loss. The three-seed main comparison uses seven configurations: the FP32 explicit model, a quantization-aware Explicit-W8A8, DEQ-Anderson, DEQ-Anderson-W8A8 with the same re-forward quantization as \qdeq, DEQ-Broyden, \qdeq-FP32 and our \qdeq-INT8. Training and the main evaluation use SA as the QUBO backend, and results are averaged over seeds 2023, 42 and 2024. Diagnostics and ablations use the seed-42 checkpoints on 256 evenly spaced test windows. The CIM experiments run on the Kaiwu CIM service with a fixed checkpoint per dataset. Additional budgeted solver swaps and consistency students use seed-42 teacher checkpoints (\cref{sec:computation,app:continuation}).

\section{Results}
\label{sec:results}

\begin{table}[!htbp]
\caption{Test MSE, mean $\pm$ std over three seeds at horizon 96. The last two columns give the relative MSE difference of \qdeq-INT8 from the Anderson and Broyden DEQs; negative values favour \qdeq. \qdeq-INT8 uses an FP32 solve and a terminal W8A8 re-forward. The best mean in each row is in bold; MAE is reported in \cref{app:tables}.}
\label{tab:main}

\centering
\small
\setlength{\tabcolsep}{4.2pt}
\begin{adjustbox}{max width=\textwidth}%
\begin{tabular}{lccccccc|cc}
\toprule
 & \multicolumn{2}{c}{Explicit} & \multicolumn{3}{c}{DEQ, continuous solver} & \multicolumn{2}{c|}{\qdeq{} (ours)} & \multicolumn{2}{c}{$\Delta$ \qdeq-INT8 (\%)} \\
\cmidrule(lr){2-3}\cmidrule(lr){4-6}\cmidrule(lr){7-8}\cmidrule(lr){9-10}
Dataset & FP32 & W8A8 & Anderson & Anderson-W8A8 & Broyden & FP32 & INT8 & vs Anderson & vs Broyden \\
\midrule
Weather & .1754{\scriptsize$\pm$.0005} & .1754{\scriptsize$\pm$.0006} & .1773{\scriptsize$\pm$.0022} & .1791{\scriptsize$\pm$.0032} & .1763{\scriptsize$\pm$.0010} & \best{.1736}{\scriptsize$\pm$.0011} & .1753{\scriptsize$\pm$.0004} & $-1.08$ & $-0.53$ \\
ETTh1 & \best{.3866}{\scriptsize$\pm$.0009} & \best{.3866}{\scriptsize$\pm$.0009} & .3940{\scriptsize$\pm$.0043} & .3940{\scriptsize$\pm$.0043} & .3948{\scriptsize$\pm$.0054} & .3978{\scriptsize$\pm$.0034} & .3978{\scriptsize$\pm$.0035} & $+0.97$ & $+0.78$ \\
ETTh2 & .3013{\scriptsize$\pm$.0004} & .3013{\scriptsize$\pm$.0004} & .2982{\scriptsize$\pm$.0007} & .2982{\scriptsize$\pm$.0008} & .2981{\scriptsize$\pm$.0005} & \best{.2978}{\scriptsize$\pm$.0001} & \best{.2978}{\scriptsize$\pm$.0001} & $-0.12$ & $-0.08$ \\
ECL & \best{.1480}{\scriptsize$\pm$.0004} & .1481{\scriptsize$\pm$.0002} & .1509{\scriptsize$\pm$.0004} & .1521{\scriptsize$\pm$.0009} & .1496{\scriptsize$\pm$.0002} & .1489{\scriptsize$\pm$.0002} & .1490{\scriptsize$\pm$.0003} & $-1.24$ & $-0.39$ \\
Traffic & \best{.3930}{\scriptsize$\pm$.0009} & .3932{\scriptsize$\pm$.0008} & .3994{\scriptsize$\pm$.0012} & .3996{\scriptsize$\pm$.0007} & .3971{\scriptsize$\pm$.0009} & .3978{\scriptsize$\pm$.0005} & .3986{\scriptsize$\pm$.0018} & $-0.21$ & $+0.37$ \\
\bottomrule
\end{tabular}\end{adjustbox}

\end{table}

\subsection{Forecasting Accuracy with SA}
\label{sec:accuracy}

\Cref{tab:main} reports accuracy with SA as the QUBO backend. \qdeq-INT8 stays within $-1.16\%$ to $+2.92\%$ of the FP32 explicit models, matches them on Weather and improves on them for ETTh2. Among the DEQ variants, it reduces MSE relative to Anderson on Weather, ETTh2, ECL and Traffic and matches Broyden, the strongest continuous solver, within $-0.53\%$ to $+0.78\%$. Quantizing the re-forward pass to W8A8 costs at most 1\% relative to \qdeq-FP32, and quantization-aware training is similarly lossless for the explicit and Anderson models, supporting the terminal W8A8 re-forward in the evaluated configurations.

\subsection{Logical Weight Storage}
\label{sec:storage}

\begin{table}[!htbp]
\caption{Logical static weight storage in MiB, counting quantized weights at one byte and all other parameters at four. This representation-based count excludes solver states, activation buffers and FP32 working copies used by the fake-quantized implementation.}
\label{tab:storage}

\centering
\small
\setlength{\tabcolsep}{3.4pt}
\begin{tabular}{lcccccc}
\toprule
 & \multicolumn{2}{c}{Explicit} & \multicolumn{2}{c}{DEQ} & \multicolumn{2}{c}{Reduction} \\
\cmidrule(lr){2-3}\cmidrule(lr){4-5}\cmidrule(lr){6-7}
Dataset & FP32 & W8A8 & FP32 & W8A8 & vs FP32 & vs W8A8 \\
\midrule
Weather & 18.44 & 4.94 & 6.40 & 1.91 & $9.7\times$ & $2.6\times$ \\
ETTh1 & 3.21 & 0.96 & 1.70 & 0.58 & $5.6\times$ & $1.7\times$ \\
ETTh2 & 0.86 & 0.29 & 0.48 & 0.20 & $4.4\times$ & $1.5\times$ \\
ECL & 18.44 & 4.94 & 6.40 & 1.91 & $9.7\times$ & $2.6\times$ \\
Traffic & 24.46 & 6.46 & 6.40 & 1.91 & $12.8\times$ & $3.4\times$ \\
\bottomrule
\end{tabular}

\end{table}

Weight tying reduces parameter count by $1.80$--$3.82\times$. Combining sharing with the eight-bit weight representation gives the logical storage in \cref{tab:storage}: $4.4$--$12.8\times$ lower than explicit FP32 and $1.5$--$3.4\times$ lower than explicit W8A8. Savings increase with explicit depth, reaching $12.8\times$ on the four-layer Traffic model. Runtime peak memory, including the FP32 tensors retained for the solve and fake quantization, is reported separately in \cref{tab:runtime}.

\subsection{Solver Behaviour}
\label{sec:diag}

\begin{figure}[!htbp]
\centering
\includegraphics[width=\textwidth,height=0.68\textheight,keepaspectratio]{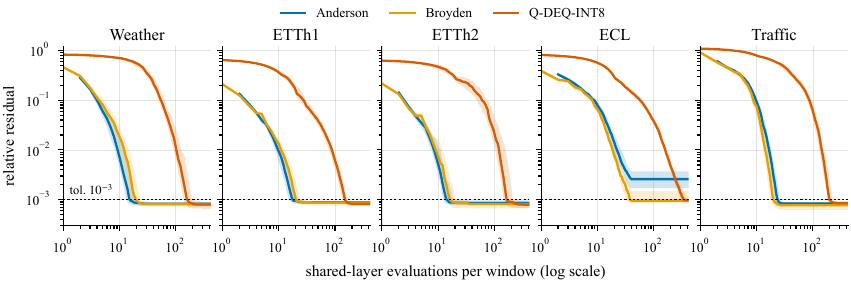}

\caption{Forward convergence. Lines show the median and bands the interquartile range over 64 test windows of the lowest relative residual reached so far, plotted against cumulative shared-layer evaluations with a common cap of 40 solver steps; the dashed line marks the tolerance $10^{-3}$.}
\label{fig:convergence}

\end{figure}

\textbf{Convergence.} \Cref{fig:convergence} traces progress against shared-layer evaluations. \qdeq{} reaches the tolerance on 73--100\% of windows, including ECL, where Anderson plateaus near $3\times10^{-3}$. \Cref{tab:diag} additionally reports both residual convergence and forecast error when continuous-solver caps increase from their default 15--25 to 40 steps. These complementary metrics distinguish equilibrium accuracy from prediction quality; query-efficient direction selection is studied in \cref{sec:computation}.

\begin{figure}[!htbp]
\centering
\includegraphics[width=0.52\columnwidth,height=0.68\textheight,keepaspectratio]{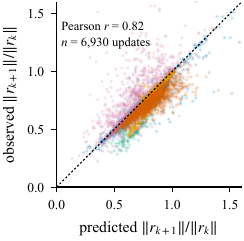}\hfill
\begin{minipage}[b]{0.45\columnwidth}
\caption{Fidelity of the local model. Residual reduction predicted by $E(a^\star)$ against the reduction observed after the step, for all \qdeq-INT8 updates on 64 windows per dataset; colours follow \cref{fig:cim}.}
\label{fig:fidelity}
\end{minipage}

\end{figure}

\textbf{Fidelity of the local model.} The quadratic objective tracks the nonlinear residual reduction: predicted and observed ratios correlate with Pearson $r=0.82$ over 6{,}930 updates and 0.69--0.96 per dataset (\cref{fig:fidelity}). The near-diagonal trend supports its use for discrete step selection, with slight optimism consistent with first-order approximation. At most 0.7\% of coefficients reach the encoded boundary under the default settings.

\begin{figure}[!htbp]
\centering
\includegraphics[width=0.86\textwidth]{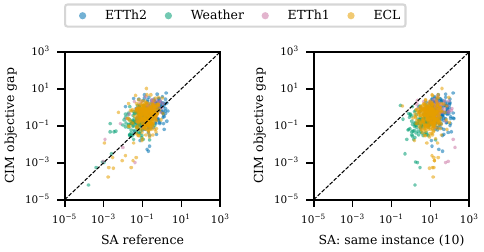}

\caption{Objective comparison on 666 archived instances from Weather, ETTh1, ETTh2 and ECL, scoring decoded solutions in the original local quadratic. Left: the configured SA reference. Right: ten returned samples on the identical submitted Ising instance. Both axes use the same continuous reference; values below $10^{-5}$ are displayed at the axis floor. Forecast metrics for all five datasets appear in \cref{tab:cim}.}
\label{fig:cim}

\end{figure}

\subsection{Encoding and Direction Selection}
\label{sec:ablation}

\begin{figure}[!htbp]
\centering
\includegraphics[width=0.63\textwidth]{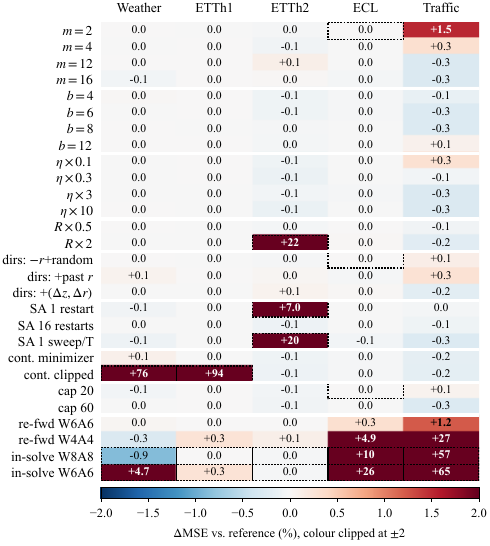}

\caption{Inference-time ablations of the seed-42 \qdeq-INT8 checkpoints on 256 paired windows, shown as relative MSE change from the reference with $m=8$, $b=10$, default $\eta$ and $R$, and SA with four restarts. Dashed boxes mark settings in which fewer than half of the windows reach the tolerance. \Cref{app:tables} lists the full results.}
\label{fig:ablation}

\end{figure}

\Cref{fig:ablation} compares inference-time changes to the seed-42 checkpoints on 256 paired test windows per dataset. Each row changes one setting from the reference $m=8$, $b=10$ configuration. Cell values report relative forecast MSE; dashed outlines identify settings where fewer than half the windows reach the residual tolerance. This separates forecast quality from fixed-point accuracy.

The upper block shows broad forecast stability across direction count, coefficient resolution and probe scale. For $m=2$--16, $b=4$--12 and $\eta/\eta_0=0.1$--10, most MSE changes lie within $\pm0.3\%$. History-based directions support convergence, and four SA restarts capture most of the forecast benefit of sixteen. The continuous-coefficient control changes MSE by only $-0.19\%$ to $+0.11\%$, supporting the central result that a finite coefficient code can preserve prediction quality.

The range $R$ determines how far the local model is trusted: doubling it raises ETTh2 MSE by approximately $22\%$, while Traffic with two directions increases MSE by $1.5\%$. We therefore select $R$ by validation and retain the coupled quadratic objective rather than clipping continuous coefficients coordinatewise. Full direction, search-budget, damping and safeguard controls appear in \cref{app:tables,app:followup,app:round3}.

\subsection{Numerical Precision}
\label{sec:precision}

\qdeq{} uses precision according to each stage's role: binary coefficients encode the update, FP32 evaluations resolve the finite-difference residual model, and W8A8 produces the terminal representation. Terminal quantization changes MSE by less than $1\%$ in the main three-seed comparison. The bitwidth sweep in \cref{fig:ablation} independently demonstrates stable forecasting with low-resolution coefficient codes.

Paired controls identify activation rounding as the main source of probe perturbations when quantization is applied throughout the solve. Weight-only W8A32 remains close to the reference forecast error, whereas in-solve W8A8 degrades ECL and Traffic; \cref{tab:partial-quant,app:tables} gives the full accuracy and convergence results. These observations motivate the mixed-precision allocation used in \qdeq{}.

\subsection{Execution on a Coherent Ising Machine}
\label{sec:cim}

Physical CIM execution supplies the coefficient decisions throughout forecasting inference on five datasets. We use the original frozen checkpoints and the same first 128 test windows as the SA references (\cref{tab:cim}). The host evaluates the network and constructs the local objective; the CIM returns ten encoded configurations per task. Device evolution time is read directly from the SDK's \texttt{time\_consume} field in microseconds.

\begin{table}[!htbp]
\caption{Original SA reference forecasts, CIM hardware forecasts, and measured device time from successful SDK task records. Task time is the mean recorded device evolution time, \texttt{time\_consume} in $\mu$s; window time is the sum over completed tasks divided by 128 windows, at the original batch settings (32 for Weather/ETT and 16 for ECL/Traffic). Each task returns ten configurations.}
\label{tab:cim}

\centering\small
\begin{adjustbox}{max width=\textwidth}%
\begin{tabular}{lrrrr}
\toprule
& \multicolumn{2}{c}{MSE} & \multicolumn{2}{c}{CIM time ($\mu$s)} \\
\cmidrule(lr){2-3}\cmidrule(lr){4-5}
Dataset & SA ref. & CIM & per task & per window \\
\midrule
Weather & 0.2568 & 0.2568 & 3318.9 & 2878.1 \\
ETTh1 & 0.2873 & 0.2873 & 3787.8 & 2426.5 \\
ETTh2 & 0.0896 & 0.0893 & 4582.1 & 5369.7 \\
ECL & 0.1123 & 0.1123 & 4713.1 & 11782.6 \\
Traffic & 0.4094 & 0.4090 & 4577.8 & 9763.7 \\
\bottomrule
\end{tabular}\end{adjustbox}
\end{table}

Across all five datasets, CIM-driven forecast MSE differs from the SA references by at most 0.348\%. Mean recorded device time is 3.3--4.7\,ms per task, with ten configurations returned per task. On the separate set of 666 archived QUBOs, CIM matches or improves the paired SA objective on 98.9--100\% of identical submitted instances (\cref{fig:cim}). The agreement at both the objective and forecasting levels validates the discrete update interface on physical hardware. \Cref{app:measured-cim} reports task counts, dispersion and MAE.

Hardware profiles retain the recorded settings for Weather, ETTh1, ETTh2 and ECL. Traffic uses uniform coupling scaling on the original 81-spin instance, decrease-only acceptance with Picard fallback, and best-residual return. This profile is selected on 16 fixed validation windows with the network checkpoint unchanged; \cref{app:traffic-profile} provides its matched-policy SA control.

\begin{figure}[!htbp]
\centering
\includegraphics[width=0.72\textwidth]{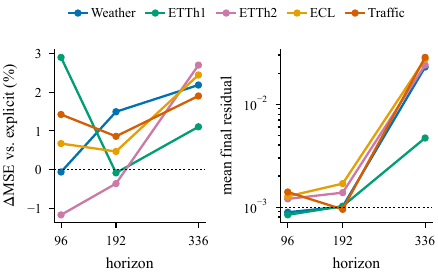}

\caption{Forecast horizons 96, 192 and 336, with one model per horizon and the three-seed mean at 96. Left: MSE gap of \qdeq-INT8 to the explicit model. Right: mean final relative residual of the forward solve.}
\label{fig:horizon}

\end{figure}

\subsection{Longer Horizons}
\label{sec:horizon}

\qdeq{} retains comparable forecasting accuracy as the horizon increases (\cref{fig:horizon}): the average MSE gap to the explicit model is about 0.5\% at horizons 96 and 192 and 2.1\% at 336. Median final residual remains below $10^{-3}$ at every horizon; a small tail of windows at 336 raises the mean.

\subsection{Computation and Accuracy}
\label{sec:computation}

\textbf{Adaptive subspaces.} The compact decision interface also permits reducing the work needed to construct it. We retain the residual direction and greedily select nonredundant directions from the eight-direction history subspace. On 256 paired windows per dataset, selecting at most four directions reduces network evaluations by 37.7--49.9\%, with relative forecast MSE changes between $-0.104\%$ and $+0.011\%$ (\cref{tab:adaptive-main}). Both methods use the same frozen checkpoint and 40-step cap. This result shows that fewer probes can retain forecast quality. Residual attainment remains a separate diagnostic: on ECL it changes from 74.2\% to 62.1\% (\cref{tab:adaptive-large}).

\begin{table}[!htbp]\centering\small
\caption{Adaptive direction selection on 256 paired windows per dataset. Queries count all shared-layer evaluations; savings and relative MSE changes are measured against the eight-direction reference.}
\label{tab:adaptive-main}
\setlength{\tabcolsep}{3.8pt}
\begin{tabular}{lrrrr}\toprule
& \multicolumn{2}{c}{Mean queries} & Saved & $\Delta$MSE \\
\cmidrule(lr){2-3}
Dataset & Fixed & Adaptive & (\%) & (\%) \\\midrule
Weather & 181.9 & 104.6 & 42.5 & -0.006 \\
ETTh1 & 156.0 & 93.8 & 39.9 & +0.011 \\
ETTh2 & 189.9 & 95.2 & 49.9 & -0.104 \\
ECL & 351.3 & 219.0 & 37.7 & +0.006 \\
Traffic & 210.7 & 131.0 & 37.8 & -0.005 \\
\bottomrule\end{tabular}\end{table}

\textbf{Composition with learned initialization.} A forecasting adaptation of C-DEQ \citep{lin2026consistency}, distilled on 1,024 training windows with validation-selected depth, supplies a learned warm start for discrete refinement. We compare this combination, standalone students and continuous solvers under 25/30-call ceilings on the same frozen teacher maps. Adaptive directions and learned initialization improve Q-DEQ's low-budget forecasts; continuous controls remain stronger on ECL and Traffic at these caps. \Cref{app:continuation} reports every setting, actual query counts, residual convergence and the additional student training and parameters.

\textbf{Execution cost.} Constructing the residual model requires $m$ probes per step, followed by the small Gram and encoding operations. The default 80-variable QUBO fixes the backend problem size across datasets. \Cref{fig:computation,app:followup} reports host/SA computation and its measured breakdown; \cref{sec:cim} reports physical device times. These measurements identify probe selection and backend integration as complementary optimization targets.

\section{Discussion}
\label{sec:discussion}

\textbf{A discrete interface for implicit computation.} \qdeq{} separates the continuous equilibrium representation from the finite coefficient decision that advances it. This creates a reusable interface: model width and task determine the residual evaluations, while direction count and bitwidth determine the QUBO. Continuous-coefficient controls and physical execution support the same finding---encoded decisions can preserve prediction quality. Adaptive directions and learned warm starts offer two compatible ways to reduce the work surrounding this interface.

\textbf{Scope and deployment.} The residual-descent analysis gives local sufficient conditions, with a separate safeguard guarantee for contractive maps. The implementation uses FP32 residual evaluations and terminal W8A8 fake quantization. Reported logical storage and CIM device times characterize their respective components; runtime memory and system latency also depend on FP32 working tensors, host computation and service communication. The prototype's SA runtime and tight-budget results (\cref{app:followup,app:continuation}) motivate integrated hardware execution and more selective probes.

\section{Conclusion}

\qdeq{} establishes a compact, binary-encoded update interface for deep equilibrium models. A local residual objective makes coefficient decisions executable as small QUBOs, preserving forecasting accuracy relative to continuous solvers and on a physical CIM. The error decomposition explains how coefficient resolution, local modelling and backend quality govern the update. Adaptive subspaces reduce network evaluations by 37.7--49.9\% with relative MSE changes within $-0.104\%$ to $+0.011\%$, while weight tying and terminal W8A8 provide complementary logical storage savings. These results support heterogeneous implicit computation in which a neural processor evaluates the representation and an Ising processor selects its update.

\section*{Acknowledgment}
ChatGPT (OpenAI) assisted with manuscript drafting, language revision and organization. The authors retain responsibility for the final text, scientific claims and reported results.

\begingroup\small
\bibliography{references}
\bibliographystyle{plainnat}
\endgroup

\clearpage
\appendix

\section{Hyperparameters}
\label{app:hparams}

\begin{table}[H]
\caption{Per-dataset solver settings. All \qdeq{} runs use $m=8$, $b=10$, history 4, mixed-history directions, tolerance $10^{-3}$, cap 40 and Kaiwu SA with initial temperature 100, geometric cooling 0.95, cutoff 0.1, five sweeps per temperature and four restarts. On Weather, $R$ is reset at every step to $1.6$ times the largest unconstrained optimal coefficient in units of $s$, clipped to $[0.5,40]$. On ETTh2, $\eta$ is set to $0.45Rs$ by the secant probe. The backward pass uses Anderson on the adjoint equation for all DEQs.}
\label{tab:hparams}

\centering
\small
\begin{adjustbox}{max width=\textwidth}
\begin{tabular}{lcccccccc}
\toprule
Dataset & $d_{\mathrm{model}}$ & batch & lr & \qdeq{} $\eta$ & \qdeq{} $R$ & Anderson $m_A$ / cap & Broyden cap & backward cap \\
\midrule
Weather & 512 & 32 & $2.07\times10^{-4}$ & 0.3 & dynamic & 8 / 25 & 25 & 35 \\
ETTh1 & 256 & 32 & $2.63\times10^{-4}$ & 0.3 & 16 & 6 / 15 & 15 & 25 \\
ETTh2 & 128 & 32 & $4.15\times10^{-5}$ & secant & 18 & 8 / 20 & 20 & 30 \\
ECL & 512 & 16 & $5\times10^{-4}$ & 0.8 & 5 & 6 / 20 & 20 & 30 \\
Traffic & 512 & 16 & $1.40\times10^{-3}$ & 0.3 & 5 & 10 / 20 & 20 & 30 \\
\bottomrule
\end{tabular}
\end{adjustbox}
\end{table}

\section{Additional Results}
\label{app:tables}

\begin{table}[H]
\caption{Test MAE, mean $\pm$ std over three seeds at horizon 96.}

\centering
\small
\begin{adjustbox}{max width=\textwidth}
\begin{tabular}{lccccccc}
\toprule
Dataset & Explicit & Explicit-W8A8 & Anderson & Anderson-W8A8 & Broyden & \qdeq-FP32 & \qdeq-INT8 \\
\midrule
Weather & .2157{\scriptsize$\pm$.0005} & .2157{\scriptsize$\pm$.0004} & .2165{\scriptsize$\pm$.0020} & .2184{\scriptsize$\pm$.0034} & .2162{\scriptsize$\pm$.0010} & .2135{\scriptsize$\pm$.0004} & .2155{\scriptsize$\pm$.0004} \\
ETTh1 & .4047{\scriptsize$\pm$.0003} & .4047{\scriptsize$\pm$.0003} & .4084{\scriptsize$\pm$.0029} & .4084{\scriptsize$\pm$.0029} & .4088{\scriptsize$\pm$.0037} & .4104{\scriptsize$\pm$.0021} & .4105{\scriptsize$\pm$.0022} \\
ETTh2 & .3508{\scriptsize$\pm$.0009} & .3508{\scriptsize$\pm$.0009} & .3494{\scriptsize$\pm$.0003} & .3494{\scriptsize$\pm$.0003} & .3496{\scriptsize$\pm$.0001} & .3499{\scriptsize$\pm$.0000} & .3499{\scriptsize$\pm$.0000} \\
ECL & .2397{\scriptsize$\pm$.0004} & .2398{\scriptsize$\pm$.0003} & .2424{\scriptsize$\pm$.0004} & .2439{\scriptsize$\pm$.0007} & .2410{\scriptsize$\pm$.0001} & .2399{\scriptsize$\pm$.0003} & .2401{\scriptsize$\pm$.0003} \\
Traffic & .2687{\scriptsize$\pm$.0007} & .2687{\scriptsize$\pm$.0003} & .2712{\scriptsize$\pm$.0008} & .2716{\scriptsize$\pm$.0005} & .2700{\scriptsize$\pm$.0005} & .2701{\scriptsize$\pm$.0003} & .2709{\scriptsize$\pm$.0008} \\
\bottomrule
\end{tabular}
\end{adjustbox}
\end{table}

\begingroup\fontsize{8.5}{10}\selectfont\setlength{\tabcolsep}{3.5pt}
\setlength{\LTpre}{7pt}\setlength{\LTpost}{7pt}
\begin{longtable}{llccccc}
\caption{Forward-solver diagnostics on 256 windows. Steps counts outer iterations for \qdeq{} and gives the index of the returned lowest-residual iterate for Anderson and Broyden. Converged is the fraction of windows reaching the tolerance $10^{-3}$. Rows with cap 40 rerun the trained Anderson and Broyden models with the step budget of \qdeq.}\label{tab:diag}\\

\toprule
Dataset & Solver & Steps & Converged & Final residual (median) & Layer evals & MSE (subset) \\
\midrule\endfirsthead

\toprule
Dataset & Solver & Steps & Converged & Final residual (median) & Layer evals & MSE (subset) \\
\midrule\endhead
\midrule\multicolumn{7}{r}{\footnotesize Continued on next page}\\\endfoot
\bottomrule\endlastfoot

Weather & Anderson (cap 25) & 16.8 & 0.89 & $8.3\times10^{-4}$ & 19 & 0.1773 \\
 & Anderson (cap 40) & 17.5 & 0.96 & $8.2\times10^{-4}$ & 20 & 0.1774 \\
 & Broyden (cap 25) & 19.6 & 0.80 & $8.8\times10^{-4}$ & 22 & 0.1782 \\
 & Broyden (cap 40) & 20.9 & 0.98 & $8.4\times10^{-4}$ & 23 & 0.1783 \\
 & Q-DEQ-INT8 (cap 40) & 20.2 & 0.99 & $8.0\times10^{-4}$ & 185 & 0.1756 \\
\midrule
ETTh1 & Anderson (cap 15) & 13.9 & 0.11 & $3.0\times10^{-3}$ & 16 & 0.3959 \\
 & Anderson (cap 40) & 20.2 & 0.95 & $8.8\times10^{-4}$ & 23 & 0.3960 \\
 & Broyden (cap 15) & 14.7 & 0.05 & $3.8\times10^{-3}$ & 17 & 0.3995 \\
 & Broyden (cap 40) & 22.1 & 0.99 & $8.6\times10^{-4}$ & 24 & 0.3995 \\
 & Q-DEQ-INT8 (cap 40) & 17.5 & 1.00 & $8.2\times10^{-4}$ & 158 & 0.3978 \\
\midrule
ETTh2 & Anderson (cap 20) & 15.0 & 0.70 & $8.9\times10^{-4}$ & 18 & 0.2960 \\
 & Anderson (cap 40) & 17.8 & 0.86 & $8.5\times10^{-4}$ & 21 & 0.2962 \\
 & Broyden (cap 20) & 16.4 & 0.73 & $8.8\times10^{-4}$ & 19 & 0.2968 \\
 & Broyden (cap 40) & 18.8 & 0.96 & $8.3\times10^{-4}$ & 21 & 0.2968 \\
 & Q-DEQ-INT8 (cap 40) & 21.6 & 0.92 & $7.9\times10^{-4}$ & 200 & 0.2964 \\
\midrule
ECL & Anderson (cap 20) & 19.0 & 0.00 & $1.5\times10^{-2}$ & 21 & 0.1513 \\
 & Anderson (cap 40) & 38.4 & 0.05 & $2.3\times10^{-3}$ & 41 & 0.1516 \\
 & Broyden (cap 20) & 19.6 & 0.00 & $7.9\times10^{-3}$ & 22 & 0.1512 \\
 & Broyden (cap 40) & 35.8 & 0.54 & $9.8\times10^{-4}$ & 39 & 0.1542 \\
 & Q-DEQ-INT8 (cap 40) & 36.5 & 0.73 & $9.7\times10^{-4}$ & 351 & 0.1490 \\
\midrule
Traffic & Anderson (cap 20) & 18.8 & 0.10 & $4.0\times10^{-3}$ & 21 & 0.4001 \\
 & Anderson (cap 40) & 24.3 & 0.96 & $8.5\times10^{-4}$ & 27 & 0.3992 \\
 & Broyden (cap 20) & 19.1 & 0.56 & $9.6\times10^{-4}$ & 21 & 0.3957 \\
 & Broyden (cap 40) & 20.8 & 0.99 & $7.6\times10^{-4}$ & 23 & 0.3946 \\
 & Q-DEQ-INT8 (cap 40) & 22.8 & 0.98 & $8.3\times10^{-4}$ & 211 & 0.3990 \\
\end{longtable}\endgroup

\begingroup\fontsize{8.5}{10}\selectfont\setlength{\tabcolsep}{2pt}
\setlength{\LTpre}{7pt}\setlength{\LTpost}{7pt}
\begin{longtable}{p{0.21\textwidth}ccccc}
\caption{Full ablation results for the seed-42 \qdeq-INT8 checkpoints on 256 paired windows. The first row gives the reference MSE and the other rows the relative MSE change in percent, followed by the fraction of converged windows and the layer evaluations per window. $n$ denotes the number of QUBO variables.}\label{layout:ablation-full}\\

\toprule
Setting & Weather & ETTh1 & ETTh2 & ECL & Traffic \\
\midrule\endfirsthead

\toprule
Setting & Weather & ETTh1 & ETTh2 & ECL & Traffic \\
\midrule\endhead
\midrule\multicolumn{6}{r}{\footnotesize Continued on next page}\\\endfoot
\bottomrule\endlastfoot

Reference ($m{=}8$, $b{=}10$, default $\eta$, $R$) & 0.1756 (0.99, 185) & 0.3978 (1.00, 158) & 0.2964 (0.92, 200) & 0.1490 (0.73, 351) & 0.3990 (0.98, 211) \\
\midrule
$m=2$ ($n=20$) & $+0.01$ (0.98, 78) & $-0.01$ (1.00, 68) & $-0.02$ (0.93, 68) & $+0.01$ (0.18, 160) & $+1.50$ (0.93, 104) \\
$m=4$ ($n=40$) & $-0.00$ (0.98, 106) & $+0.00$ (1.00, 93) & $-0.07$ (0.93, 103) & $-0.01$ (0.54, 224) & $+0.32$ (0.96, 133) \\
$m=12$ ($n=120$) & $+0.01$ (0.95, 309) & $+0.00$ (1.00, 244) & $+0.12$ (0.92, 338) & $-0.01$ (0.74, 487) & $-0.29$ (0.99, 300) \\
$m=16$ ($n=160$) & $-0.06$ (0.89, 424) & $+0.01$ (1.00, 340) & $+0.03$ (0.92, 445) & $-0.04$ (0.70, 630) & $-0.29$ (0.99, 377) \\
\midrule
$b=4$ ($n=32$) & $+0.03$ (0.98, 195) & $+0.00$ (1.00, 184) & $-0.09$ (0.93, 229) & $-0.02$ (0.75, 350) & $-0.14$ (0.98, 212) \\
$b=6$ ($n=48$) & $-0.00$ (0.99, 184) & $+0.00$ (1.00, 158) & $-0.08$ (0.95, 190) & $-0.02$ (0.74, 349) & $-0.28$ (0.99, 210) \\
$b=8$ ($n=64$) & $+0.02$ (0.99, 184) & $-0.01$ (1.00, 155) & $-0.03$ (0.94, 193) & $-0.01$ (0.75, 350) & $-0.27$ (0.98, 210) \\
$b=12$ ($n=96$) & $+0.01$ (0.99, 187) & $+0.01$ (1.00, 160) & $-0.04$ (0.93, 202) & $-0.01$ (0.73, 352) & $+0.08$ (0.97, 215) \\
\midrule
$\eta\times0.1$ & $-0.00$ (0.99, 172) & $+0.00$ (1.00, 157) & $-0.08$ (0.91, 186) & $-0.01$ (0.61, 365) & $+0.31$ (0.64, 355) \\
$\eta\times0.3$ & $-0.01$ (0.99, 174) & $+0.00$ (1.00, 157) & $-0.12$ (0.93, 192) & $-0.00$ (0.70, 355) & $-0.05$ (0.96, 240) \\
$\eta\times3$ & $-0.00$ (0.98, 201) & $-0.01$ (1.00, 158) & $-0.06$ (0.96, 215) & $-0.02$ (0.73, 352) & $-0.29$ (0.98, 204) \\
$\eta\times10$ & $+0.00$ (0.98, 198) & $-0.01$ (1.00, 168) & $-0.13$ (0.91, 269) & $-0.01$ (0.66, 355) & $-0.29$ (0.98, 207) \\
\midrule
$R\times0.5$ & $+0.01$ (0.99, 174) & $-0.01$ (1.00, 150) & $-0.04$ (0.91, 174) & $+0.01$ (0.72, 352) & $-0.07$ (0.98, 203) \\
$R\times2$ & $-0.01$ (0.98, 221) & $-0.01$ (0.99, 219) & $+21.95$ (0.02, 401) & $+0.02$ (0.61, 366) & $-0.22$ (0.95, 262) \\
\midrule
Directions: $-r$ + random only & $+0.00$ (0.94, 225) & $-0.01$ (1.00, 188) & $-0.03$ (0.88, 210) & $+0.01$ (0.00, 402) & $+0.13$ (0.92, 276) \\
Directions: + past residuals & $+0.07$ (0.98, 175) & $+0.00$ (1.00, 157) & $+0.04$ (0.89, 179) & $-0.02$ (0.63, 362) & $+0.28$ (0.96, 218) \\
Directions: + $(\Delta z,\Delta r)$ pairs & $-0.01$ (0.99, 183) & $-0.00$ (1.00, 175) & $+0.07$ (0.92, 222) & $+0.02$ (0.68, 356) & $-0.22$ (0.98, 216) \\
\midrule
SA: 1 restart & $-0.07$ (0.93, 233) & $-0.01$ (1.00, 215) & $+7.05$ (0.41, 355) & $-0.01$ (0.57, 371) & $-0.04$ (0.95, 246) \\
SA: 16 restarts & $-0.00$ (0.99, 174) & $-0.01$ (1.00, 150) & $-0.10$ (0.94, 180) & $-0.01$ (0.74, 348) & $-0.08$ (0.99, 203) \\
SA: 1 sweep per temperature & $-0.11$ (0.88, 258) & $-0.01$ (0.88, 302) & $+19.82$ (0.02, 401) & $-0.05$ (0.52, 373) & $-0.27$ (0.93, 272) \\
\midrule
Continuous minimizer, unconstrained & $+0.11$ (0.99, 165) & $-0.01$ (1.00, 170) & $-0.09$ (0.95, 181) & $-0.02$ (0.78, 343) & $-0.19$ (0.99, 192) \\
Continuous minimizer, clipped to code box & $+76.26$ (0.00, 402) & $+94.32$ (0.00, 402) & $-0.10$ (0.89, 223) & $-0.02$ (0.76, 343) & $-0.19$ (0.99, 192) \\
\midrule
Step cap 20 & $-0.06$ (0.75, 170) & $+0.00$ (0.97, 156) & $-0.06$ (0.74, 174) & $+0.00$ (0.00, 202) & $+0.07$ (0.58, 192) \\
Step cap 60 & $-0.01$ (0.99, 188) & $+0.00$ (1.00, 159) & $-0.07$ (0.93, 215) & $-0.01$ (0.88, 384) & $-0.27$ (0.99, 213) \\
\midrule
Re-forward FP32 & $+0.00$ (0.99, 185) & $-0.00$ (1.00, 158) & $-0.00$ (0.92, 200) & $-0.03$ (0.73, 351) & $-0.05$ (0.98, 211) \\
Re-forward W6A6 & $+0.04$ (0.99, 185) & $+0.01$ (1.00, 158) & $+0.01$ (0.92, 200) & $+0.32$ (0.73, 351) & $+1.22$ (0.98, 211) \\
Re-forward W4A4 & $-0.28$ (0.99, 185) & $+0.27$ (1.00, 158) & $+0.12$ (0.92, 200) & $+4.95$ (0.73, 351) & $+26.53$ (0.98, 211) \\
\midrule
In-solve W8A8 & $-0.87$ (0.00, 402) & $+0.00$ (0.00, 402) & $-0.02$ (0.00, 402) & $+10.28$ (0.00, 402) & $+56.78$ (0.00, 402) \\
In-solve W6A6 & $+4.71$ (0.00, 402) & $+0.28$ (0.00, 402) & $-0.03$ (0.00, 402) & $+26.45$ (0.00, 402) & $+65.44$ (0.00, 402) \\
\end{longtable}\endgroup

\begin{table}[H]
\caption{Quantizing inside the solve. Test MSE on 256 windows, followed by the fraction of windows that reach the tolerance.}
\label{tab:insolve}

\centering
\small
\begin{adjustbox}{max width=\textwidth}
\begin{tabular}{lcccc}
\toprule
 & \multicolumn{2}{c}{Anderson-W8A8} & \multicolumn{2}{c}{\qdeq-INT8} \\
\cmidrule(lr){2-3}\cmidrule(lr){4-5}
 & re-forward & in-solve & re-forward & in-solve \\
\midrule
Weather & .1762 (0.73) & .1762 (0.00) & .1756 (0.99) & .1741 (0.00) \\
ETTh1 & .3958 (0.11) & .3958 (0.00) & .3978 (1.00) & .3978 (0.00) \\
ETTh2 & .2960 (0.70) & .2960 (0.00) & .2964 (0.92) & .2963 (0.00) \\
ECL & .1512 (0.00) & .1509 (0.00) & .1490 (0.73) & .1643 (0.00) \\
Traffic & .3989 (0.06) & .3987 (0.00) & .3990 (0.98) & .6256 (0.00) \\
\bottomrule
\end{tabular}
\end{adjustbox}
\end{table}

\begingroup\fontsize{8.5}{10}\selectfont\setlength{\tabcolsep}{2.5pt}
\setlength{\LTpre}{7pt}\setlength{\LTpost}{7pt}
\begin{longtable}{llcccccc}
\caption{Inference cost on conventional hardware at batch size 1, measured on 256 windows on an RTX 4090 GPU and 64 windows on four CPU threads with the seed-42 checkpoints; \qdeq{} uses SA on the host CPU. GPU energy is board power sampled at 20\,Hz times wall time.}\label{tab:runtime}\\

\toprule
Dataset & Configuration & GPU ms/window & Layer evals & Peak GPU MB & GPU J/window & CPU ms/window & MSE (subset) \\
\midrule\endfirsthead

\toprule
Dataset & Configuration & GPU ms/window & Layer evals & Peak GPU MB & GPU J/window & CPU ms/window & MSE (subset) \\
\midrule\endhead
\midrule\multicolumn{8}{r}{\footnotesize Continued on next page}\\\endfoot
\bottomrule\endlastfoot

Weather & Explicit & 1.6 & 3 & 28 & 0.10 & 2.4 & 0.1750 \\
 & Explicit-W8A8 & 4.4 & 3 & 30 & 0.29 & 6.4 & 0.1752 \\
 & Anderson & 16.1 & 19 & 17 & 1.34 & 19.3 & 0.1773 \\
 & Anderson-W8A8 & 19.9 & 22 & 18 & 1.65 & -- & 0.1762 \\
 & Broyden & 18.1 & 22 & 19 & 1.50 & 27.0 & 0.1782 \\
 & Q-DEQ-FP32 & 917.7 & 247 & 18 & 65.35 & -- & 0.1765 \\
 & Q-DEQ-INT8 & 690.3 & 184 & 18 & 49.12 & 737.8 & 0.1756 \\
\midrule
ETTh1 & Explicit & 1.2 & 2 & 12 & 0.08 & 1.0 & 0.3885 \\
 & Explicit-W8A8 & 3.1 & 2 & 13 & 0.19 & 3.0 & 0.3885 \\
 & Anderson & 12.9 & 16 & 11 & 1.00 & 9.4 & 0.3959 \\
 & Anderson-W8A8 & 14.0 & 16 & 11 & 1.11 & -- & 0.3958 \\
 & Broyden & 13.7 & 17 & 11 & 1.09 & 11.5 & 0.3995 \\
 & Q-DEQ-FP32 & 572.3 & 156 & 11 & 40.96 & -- & 0.3977 \\
 & Q-DEQ-INT8 & 578.1 & 157 & 11 & 41.64 & 561.6 & 0.3978 \\
\midrule
ETTh2 & Explicit & 1.2 & 2 & 10 & 0.07 & 0.8 & 0.3008 \\
 & Explicit-W8A8 & 3.1 & 2 & 10 & 0.21 & 2.0 & 0.3008 \\
 & Anderson & 14.3 & 18 & 10 & 1.19 & 8.6 & 0.2960 \\
 & Anderson-W8A8 & 15.9 & 18 & 10 & 1.31 & -- & 0.2960 \\
 & Broyden & 15.3 & 19 & 10 & 1.26 & 10.2 & 0.2968 \\
 & Q-DEQ-FP32 & 736.0 & 200 & 10 & 53.47 & -- & 0.2961 \\
 & Q-DEQ-INT8 & 733.6 & 199 & 10 & 52.68 & 686.7 & 0.2962 \\
\midrule
ECL & Explicit & 2.1 & 3 & 40 & 0.17 & 12.7 & 0.1482 \\
 & Explicit-W8A8 & 5.9 & 3 & 40 & 0.48 & 18.8 & 0.1481 \\
 & Anderson & 18.1 & 21 & 43 & 1.82 & 102.9 & 0.1513 \\
 & Anderson-W8A8 & 22.8 & 21 & 43 & 2.28 & -- & 0.1512 \\
 & Broyden & 22.5 & 22 & 61 & 2.31 & 133.4 & 0.1512 \\
 & Q-DEQ-FP32 & 1348.4 & 354 & 60 & 102.50 & -- & 0.1490 \\
 & Q-DEQ-INT8 & 1344.4 & 351 & 60 & 99.90 & 2664.5 & 0.1490 \\
\midrule
Traffic & Explicit & 2.6 & 4 & 112 & 0.25 & 85.1 & 0.3930 \\
 & Explicit-W8A8 & 6.0 & 4 & 112 & 0.72 & 95.8 & 0.3923 \\
 & Anderson & 20.0 & 21 & 152 & 4.15 & 549.3 & 0.4001 \\
 & Anderson-W8A8 & 22.8 & 21 & 152 & 4.50 & -- & 0.3988 \\
 & Broyden & 20.2 & 21 & 181 & 4.51 & 592.6 & 0.3957 \\
 & Q-DEQ-FP32 & 791.4 & 207 & 180 & 74.77 & -- & 0.3967 \\
 & Q-DEQ-INT8 & 781.4 & 210 & 180 & 73.68 & 5040.1 & 0.3979 \\
\end{longtable}\endgroup

\begin{table}[H]
\caption{Archived forecast metrics and physical-spin dimensions for the four replay cohorts underlying \cref{fig:cim}. Both backends select binary coefficient codes. The fresh five-dataset hardware run and its measured task times are reported in \cref{tab:cim,app:measured-cim}.}

\centering
\small
\begin{adjustbox}{max width=\textwidth}
\begin{tabular}{lcccc}
\toprule
Dataset & QUBOs & spins & SA MSE/MAE & CIM MSE/MAE \\
\midrule
Weather & 113 & 100--127 & .2568/.3044 & .2568/.3043 \\
ETTh1 & 88 & 105--140 & .2873/.3511 & .2873/.3511 \\
ETTh2 & 145 & 106--135 & .0896/.2095 & .0897/.2090 \\
ECL & 320 & 91--132 & .1123/.2101 & .1123/.2101 \\
\bottomrule
\end{tabular}
\end{adjustbox}
\end{table}

\section{Graph Matching}
\label{app:willow}

To check that the solver is not specific to forecasting, we inserted it into an NMT-style graph matching model whose weight-tied decoder combines self-attention, cross-attention and an MLP. We report a single run on the five-category Willow ObjectClass benchmark, with all configurations starting from the same explicit checkpoint and \qdeq{} using eight candidate directions with Kaiwu SA. Mean matching accuracy is 0.9796 for the explicit model, 0.9780 for DEQ-Anderson, 0.9743 for \qdeq-FP32 and 0.9781 for \qdeq-INT8. The Duck category, the hardest, ranges from 0.9148 to 0.9388.

\section{Additional Computational Experiments}
\label{app:followup}
\begin{figure}[!ht]\centering
\includegraphics[width=\textwidth,height=0.68\textheight,keepaspectratio]{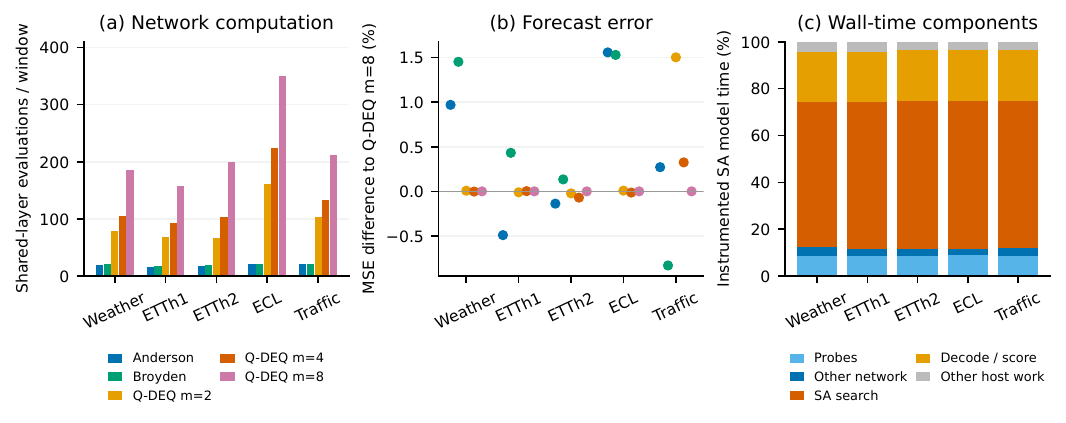}
\caption{Computation and accuracy from recorded experiments. (a,b) Network evaluations and forecast MSE on 256 windows per dataset, using native continuous-solver checkpoints and inference-time changes to the Q-DEQ checkpoint. The MSE reference is Q-DEQ with eight directions. (c) Synchronized SA runtime decomposition on a separate 32-window cohort. All runtime components in (c) belong to the recorded SA implementation.}
\label{fig:computation}\end{figure}

\paragraph{Recorded SA execution cost.} At batch one on the RTX 4090, the recorded SA implementation takes 578--1344\,ms per window, compared with 14--23\,ms for Broyden. The synchronized decomposition assigns 8.5--8.9\% of model time to probes and 83--85\% to SA search plus candidate decoding/scoring (\cref{fig:computation}). These measurements describe the host/SA implementation; device evolution times for physical CIM execution are reported in \cref{tab:cim}.

\textbf{Parallel probes.} We benchmark eight forward-difference probes on the initial DEQ states of three test windows per dataset using the seed-42 checkpoints, an RTX 4090 and PyTorch 2.4.1. Each mode receives three warm-ups and sixteen repetitions in alternating order, with GPU synchronization around every timed group. The wall time includes stream scheduling and joining. Outputs agree elementwise; 2, 4 and 8 streams do not accelerate this implementation. These measurements characterize the initial-state probe kernel rather than complete inference.
\begin{table}[H]\centering\small
\caption{Probe-group wall time (ms), median across three states.}
\begin{tabular}{lrrrr}\toprule Dataset & Serial & 2 streams & 4 streams & 8 streams \\ \midrule
weather & 3.122 & 3.441 & 3.433 & 3.476 \\
ETTh1 & 3.004 & 3.331 & 3.312 & 3.365 \\
ETTh2 & 3.017 & 3.340 & 3.335 & 3.383 \\
ECL & 3.049 & 3.362 & 3.387 & 3.403 \\
traffic & 3.046 & 3.382 & 3.396 & 3.435 \\
\bottomrule\end{tabular}\end{table}
\textbf{Safeguard.} Each run covers the same 256 windows as the reference. Picard fallback on this backbone is an empirical update rule; contractivity is not assumed.
\begin{table}[H]\centering\small
\caption{Paired safeguard results: MSE, convergence fraction and mean steps.}
\begin{adjustbox}{max width=\textwidth}\begin{tabular}{lrrrrrr}\toprule & \multicolumn{2}{c}{MSE} & \multicolumn{2}{c}{Converged} & \multicolumn{2}{c}{Steps} \\ Dataset & Base & Safe & Base & Safe & Base & Safe \\ \midrule
weather & 0.1756 & 0.1756 & 0.9883 & 0.9883 & 20.1992 & 19.4961 \\
ETTh1 & 0.3978 & 0.3978 & 0.9961 & 0.9961 & 17.4648 & 17.4648 \\
ETTh2 & 0.2964 & 0.2960 & 0.9180 & 0.9492 & 21.6016 & 20.0508 \\
ECL & 0.1490 & 0.1490 & 0.7266 & 0.7227 & 36.5117 & 36.5078 \\
traffic & 0.3990 & 0.3979 & 0.9844 & 0.9844 & 22.7969 & 22.6289 \\
\bottomrule\end{tabular}\end{adjustbox}\end{table}
\textbf{Batched production probes.} A second microbenchmark uses the production initial directions and effective adaptive finite-difference step. Packing eight probes into one network batch gives speedups of 5.99, 5.65, 5.64, 4.43 and 1.25 on Weather, ETTh1, ETTh2, ECL and Traffic. The largest relative L2 probe difference across states is approximately 3.4\%. These are probe-group measurements; the paired inference pilot below measures their effect on the complete solver.
\textbf{Paired inference pilot.} On eight evenly spaced windows per dataset, with identical checkpoints and seeds, batched probes give the following model-time speedups and MSE changes: weather: 1.203x / -0.039\% MSE; ETTh1: 1.061x / -0.022\% MSE; ETTh2: 1.129x / +0.028\% MSE; ECL: 1.066x / +0.011\% MSE; traffic: 1.011x / +0.011\% MSE. Each configuration is timed once; the sample size and timing repetitions are distinct from the 256-window main evaluation.

\subsection{Scaling and damping details}
\begin{table}[H]\centering\small
\caption{Relative MSE change (\%) on 256 paired windows per dataset.}
\begin{adjustbox}{max width=\textwidth}\begin{tabular}{lrrrrr}\toprule Setting & Weather & ETTh1 & ETTh2 & ECL & Traffic \\ \midrule
$b=14$ & +0.002 & +0.001 & -0.081 & -0.002 & -0.048 \\
$b=16$ & -0.014 & -0.010 & -0.128 & -0.001 & -0.279 \\
$m=24$ & +0.294 & +0.001 & -0.107 & -0.006 & -0.278 \\
$m=32$ & +2.641 & +83.436 & +38.460 & -0.022 & -0.286 \\
$\lambda=.03$ & +0.002 & +0.005 & -0.071 & -0.021 & -0.302 \\
$\lambda=.1$ & +0.000 & +0.003 & -0.072 & +0.005 & -0.013 \\
$\lambda=.3$ & -0.005 & -0.012 & -0.098 & +0.008 & -0.048 \\
$\lambda=1$ & -0.004 & +0.004 & -0.066 & +0.004 & -0.145 \\
Range control & -0.077 & -0.009 & -0.169 & -0.006 & -0.115 \\
\bottomrule\end{tabular}\end{adjustbox}\end{table}

\section{Encoding and Safeguarded Convergence}
\label{app:derivation}

\subsection{Local quadratic, QUBO and auxiliary spin}
Let $o=\mathrm{off}$, $a=o+B\xi$ and $\xi\in\{0,1\}^{n}$. With $G=P^\top P$ and $h=P^\top r$,
\begin{align}
E(a)&=\tfrac12 a^\top Ga+h^\top a\\
&=\xi^\top Q\xi+q^\top\xi+C,\\
Q&=\tfrac12B^\top GB,\quad q=B^\top(Go+h),\quad C=\tfrac12o^\top Go+h^\top o.
\end{align}
Writing $s=2\xi-\mathbf1$ gives $s^\top Js+\ell^\top s+C_s$, where $J_{ij}=Q_{ij}/4$ for $i\ne j$, $J_{ii}=0$,
\begin{align}
\ell&=\tfrac12(Q\mathbf1+q),\\
C_s&=C+\tfrac14\mathbf1^\top Q\mathbf1+\tfrac12q^\top\mathbf1+\tfrac14\operatorname{tr}(Q).
\end{align}
One auxiliary spin $\tau$ absorbs the field: $H(s,\tau)=s^\top Js+\tau\ell^\top s$. The augmented matrix has off-diagonal last-column entries $\ell_i/2$. Decode $\xi=(\tau s+\mathbf1)/2$; a global spin flip leaves the decoded bits unchanged. Positive energy rescaling preserves the minimizer; the service adapter handles its sign convention. Exhaustive checks over all 16 assignments of a four-bit instance and both auxiliary-spin signs reproduce the direct local objective within $1.34\times10^{-15}$.

With $v_1=-r/\|r\|$, the Picard displacement corresponds to $a_1=-\|r\|$ and all other coefficients zero. It lies in the continuous box if $R\geq\sqrt m$; finite even-level grids generally approximate rather than exactly contain that vector. The algorithm therefore evaluates Picard fallback explicitly.

\subsection{Finite differences, discrete codes and noncontractive maps}
\label{app:model-bound}
Fix an input and write $r(z)=F(z)-z$. Suppose $r$ is continuously differentiable with an $L$-Lipschitz Jacobian on a convex neighborhood containing the probe and update segments. Each column of $V\in\mathbb R^{d\times m}$ has Euclidean norm one. For $p_i=[r(z+\eta v_i)-r(z)]/\eta$, the fundamental theorem of calculus gives
\[
\|p_i-J_r(z)v_i\|\leq L\eta/2,\qquad
\|r(z+Va)-r(z)-J_r(z)Va\|\leq L\|Va\|^2/2.
\]
Consequently,
\begin{equation}
\|r(z+Va)-(r+Pa)\|\leq \frac{L\eta}{2}\|a\|_1+\frac L2\|Va\|^2=D(a).
\label{eq:defect-proof}
\end{equation}
This is a deterministic bound for the undamped local residual objective; the first inequality applies to any candidate, including a damped one.

\paragraph{Grid and backend error.} Set $s=\|r\|/\sqrt m$ and $\mathcal B=[-Rs,Rs]^m$. Let $a_\Box$ minimize $E(a)=\frac12\|r+Pa\|^2-\frac12\|r\|^2$ on $\mathcal B$, and let $a_g$ minimize it on the uniform $b$-bit grid. Rounding each coordinate of $a_\Box$ gives a feasible point $a_b$ with
\[
\|a_b-a_\Box\|\leq\frac{\sqrt m Rs}{2^b-1}
=\frac{R\|r\|}{2^b-1}.
\]
For a backend result $\widehat a$ satisfying $E(\widehat a)-E(a_g)\leq\varepsilon_Q$,
\begin{align}
\|r+P\widehat a\|
&\leq\sqrt{\|r+Pa_b\|^2+2\varepsilon_Q}\\
&\leq\|r+Pa_\Box\|+\frac{R\|P\|_2\|r\|}{2^b-1}+\sqrt{2\varepsilon_Q}.
\end{align}
Combining this with \cref{eq:defect-proof} proves \cref{eq:discrete-bound}. The backend gap is a condition on solution quality; it is not assumed to be certified by SA or CIM. The continuous reference here is the box optimum, which generally differs from clipping the unconstrained minimizer.

\paragraph{Conditional convergence for a noncontractive map.} Suppose a sequence of grid updates satisfies the bound in \cref{eq:discrete-bound} with a common right-hand side at most $\sigma\|r(z_k)\|$, for $\sigma<1$. Assume $R_k\leq\overline R$, $\|V_k\|_2\leq\overline V$ and continuity of $r$ on the closure of the visited states. Then
\[
\|r(z_k)\|\leq\sigma^k\|r(z_0)\|,\qquad
\|z_{k+1}-z_k\|\leq\overline V\overline R\sigma^k\|r(z_0)\|.
\]
The second inequality follows from $\|a_k\|\leq R_k\|r(z_k)\|$. Summing its geometric tail makes $\{z_k\}$ Cauchy in $\mathbb R^d$; its limit $z^\star$ satisfies $r(z^\star)=0$ and
\[
\|z_k-z^\star\|\leq\frac{\overline V\overline R}{1-\sigma}\sigma^k\|r(z_0)\|.
\]
These are sufficient conditions on the local model, subspace and optimization accuracy. They do not assert uniqueness of the root or contraction of $F$, and they need not hold on every forecasting window. Decrease-only acceptance with $\sigma=1$ does not imply this geometric rate. Picard fallback requires the separate contractive-map assumptions of the following proposition.

\paragraph{Bounded evaluation errors.} If each evaluated residual has error norm at most $\nu$, write $\widetilde r$ and $\widetilde P$ for the measured residual and forward differences. Directly bounding the two errors in each probe gives
\[
\|r(z+Va)\|\leq\|\widetilde r+\widetilde P a\|+\nu+
\left(\frac{L\eta}{2}+\frac{2\nu}{\eta}\right)\|a\|_1+\frac L2\|Va\|^2.
\]
Thus smaller probes reduce truncation error while amplifying bounded evaluation errors. The result requires smoothness along the evaluated segments; it makes no smoothness claim across an activation kink or a quantization discontinuity.

\paragraph{Controlled numerical checks.} We use two 16-dimensional smooth residual families,
\[
r(z)=Az+c+0.15\tanh(z),\qquad
A=\alpha I+W,\quad W^\top=-W,\quad\|W\|_2=0.12,
\]
with $\alpha=0.45$ (expansive) or $\alpha=-2.6$ (oscillatory), $c_i\sim\mathcal N(0,0.1^2)$, and $z_{0,i}\sim\mathcal N(0,0.25^2)$. The map is $F(z)=z+r(z)$. Its Jacobian norm is at least 1.45 in both families, and the residual Jacobian has Lipschitz bound $L=0.15\cdot4/(3\sqrt3)$. For each family and seeds 0--19, the subspace has the normalized negative residual and one normalized Gaussian direction. We enumerate the complete coefficient grid and solve the two-dimensional continuous box reference by enumerating its active faces.

Across $\eta\in\{0.01,0.1,0.5\}$, $R\in\{1,4\}$ and $b\in\{3,4,6,8\}$, all 960 one-step cases satisfy both the finite-difference and grid bounds. Another 240 checks with bounded residual errors $\nu\in\{10^{-6},10^{-3}\}$ and $\eta\in\{0.001,0.01,0.1\}$ satisfy the noise bound. A fixed trajectory protocol ($m=2$, $b=6$, $R=4$, $\eta=0.01$, cap 60) uses exact grid selection and no Picard fallback. All 40 runs reach absolute residual below $10^{-6}$; \cref{tab:noncontractive} summarizes the measured query counts. Each update includes the base residual, two probes and candidate validation, and the terminal residual is also counted. These controlled maps isolate the error mechanism and the sufficient conditions, separately from the trained forecasting experiments.

\begin{table}[H]\centering\small
\caption{Controlled noncontractive maps, 20 seeds per family, using exact coefficient-grid enumeration. The certificate is the largest bound on the next/current residual ratio over all executed updates.}
\label{tab:noncontractive}
\begin{tabular}{lrrrr}\toprule
Family & Mean updates & Mean queries & Max final residual & Max certificate \\\midrule
expansive & 8.00 & 33.00 & $7.01\times10^{-7}$ & 0.328 \\
oscillatory & 7.00 & 29.00 & $5.10\times10^{-7}$ & 0.246 \\
\bottomrule\end{tabular}\end{table}
\begin{figure}[!htbp]\centering
\includegraphics[width=.83\textwidth,height=0.68\textheight,keepaspectratio]{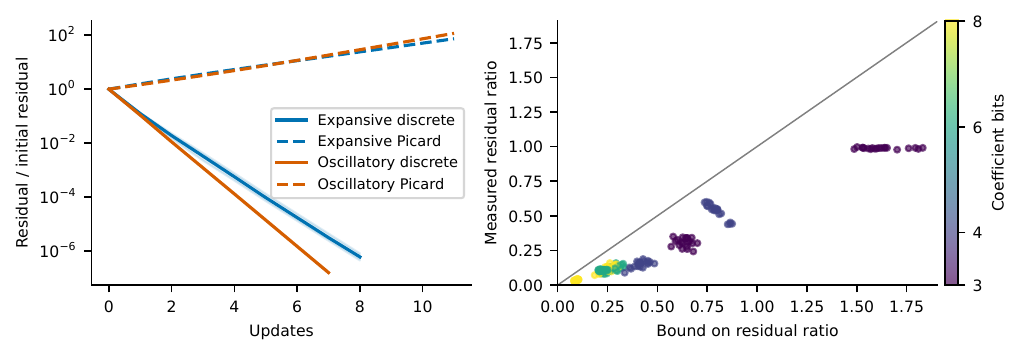}
\caption{Controlled noncontractive maps. Left: median residual ratios, with interquartile bands for discrete updates; Picard trajectories are shown for 12 residual evaluations. Right: measured and bounded one-step ratios for $\eta=0.01$, $R=4$ and all tested bitwidths. All numerical checks and trajectories are retained in the accompanying records.}
\label{fig:model-bound}\end{figure}

\paragraph{Magnitude of local-model terms on trained maps.}
\label{app:trained-terms}
We examine the archived seed-42 FP32 Q-DEQ checkpoints for ETTh1 and ETTh2 on 16 evenly spaced test inputs each. The original layer, input normalization, timestamp tokens and stored parameters are used, with dropout disabled. Initial, midpoint and returned states from an Anderson solve on the same frozen map provide three observation locations per input. At each location, the original mixed-history rule forms eight directions, with $R=16$, $\eta=0.3$ on ETTh1 and $R=18$, $\eta=\max(0.45Rs,10^{-4})$ on ETTh2. This gives 48 states per dataset. These observation locations are a common-map diagnostic rather than recorded SA/CIM trajectories.

Automatic differentiation computes $J_r(z)V$. A feasible box least-squares control is rounded to the ten-bit grid to obtain $\widehat a$, and the residual is evaluated at $z+V\widehat a$. We measure
\[
d_{\rm FD}=\|(P-J_rV)\widehat a\|,\qquad
d_{\rm NL}=\|r(z+V\widehat a)-r(z)-J_rV\widehat a\|.
\]
Their sum bounds the measured model defect $\|r(z+V\widehat a)-(r+P\widehat a)\|$ by the triangle inequality. These vector errors can cancel; their separate norms describe magnitudes rather than additive attribution. Table~\ref{tab:trained-terms} reports these observations and the normalized grid term from Equation~\ref{eq:discrete-bound}. All candidate evaluations satisfy the measured-defect inequality within FP32 numerical tolerance.

\begin{table}[H]\centering\small
\caption{Trained-map local-model observations: median (90th percentile), 16 states per row. The grid term, $d_{\rm FD}$, $d_{\rm NL}$ and total defect are divided by the current residual norm. The coefficient control is rounded box least squares; these are diagnostic values rather than SA/CIM optimality-gap measurements.}
\label{tab:trained-terms}
\begin{adjustbox}{max width=\textwidth}
\begin{tabular}{llrrrrr}\toprule
Dataset & State & $\|P\|_2$ & Grid term & $d_{\rm FD}/\|r\|$ & $d_{\rm NL}/\|r\|$ & Total defect$/\|r\|$ \\\midrule
ETTh1 & initial & 1.2214 (2.2169) & 0.0191 (0.0347) & 0.0108 (0.0248) & 0.2741 (0.2827) & 0.2850 (0.2951) \\
ETTh1 & middle & 0.9643 (1.1821) & 0.0151 (0.0185) & 0.0337 (0.0417) & 0.0755 (0.1324) & 0.0934 (0.1611) \\
ETTh1 & returned & 0.5286 (0.7148) & 0.0083 (0.0112) & 0.0418 (0.0589) & 0.0090 (0.0183) & 0.0501 (0.0736) \\
ETTh2 & initial & 0.9893 (1.0197) & 0.0174 (0.0179) & 0.9813 (1.8781) & 0.7651 (1.6699) & 0.3529 (0.4843) \\
ETTh2 & middle & 1.2469 (1.7631) & 0.0219 (0.0310) & 0.2373 (0.4685) & 0.0462 (0.1544) & 0.2763 (0.5707) \\
ETTh2 & returned & 0.6986 (1.2633) & 0.0123 (0.0222) & 0.0070 (0.0755) & 0.0093 (0.0627) & 0.0161 (0.1298) \\
\bottomrule\end{tabular}\end{adjustbox}\end{table}

The median grid term is 0.0083--0.0219 across the six groups. Finite-difference and nonlinear-model errors vary with the observation location; on ETTh2 their individual norms can be much larger than the total defect. The secant-scale probes therefore need to be assessed against the actual proposed displacement, together with the observed residual reduction. For each displacement, $2d_{\rm NL}/\|V\widehat a\|^2$ gives a directional curvature observation. These finite observations characterize sampled directions and do not certify a Jacobian Lipschitz upper bound over a neighborhood. The diagnostic uses FP32 network evaluations on a CPU and reports no inference-speed comparison.

\paragraph{A computable backend-gap upper reference.} Let $f(a)=\frac12\|r+Pa\|^2$ on $\mathcal B=[-Rs,Rs]^m$. For any feasible $a_c$, convexity gives a lower reference
\[
\ell_c=\max\{0,\ f(a_c)-\nabla f(a_c)^\top a_c-Rs\|\nabla f(a_c)\|_1\}
\leq\min_{a\in\mathcal B}f(a).
\]
For a feasible grid result $\widehat a$, the gap to the best grid point is consequently at most $f(\widehat a)-\ell_c$. This provides a conservative way to quantify the backend term without identifying the binary optimum. Its numerical tightness depends on the continuous control; verified floating-point certification would additionally require rounding-error bounds. The present trained-map diagnostic uses this control to define the test displacement; it makes no claim about the gap distribution of the recorded SA/CIM runs.

\subsection{Uniform safeguard proposition}
\paragraph{Proposition.} Let $F:D\to D$ be a contraction with constant $q<1$ on a closed complete invariant domain $D$. Set $r(z)=F(z)-z$ and fix $0<\gamma<1$. Accept a discrete candidate $z_c\in D$ only when $\|r(z_c)\|\leq\gamma\|r(z)\|$; otherwise set $z^+=F(z)$. Then $F$ has a unique fixed point $z^\star$, and
\[
\|r(z_k)\|\leq\max(q,\gamma)^k\|r(z_0)\|,\qquad
\|z_k-z^\star\|\leq\frac{\|r(z_k)\|}{1-q}.
\]
\begin{proof}
An accepted candidate satisfies the first contraction bound by construction. For fallback, $\|r(F(z))\|=\|F(F(z))-F(z)\|\leq q\|r(z)\|$. Induction gives the geometric envelope. The fixed-point error estimate follows from $(1-q)\|z-z^\star\|\leq\|F(z)-z\|$ and Banach's theorem.
\end{proof}
This proposition concerns the absolute residual. Algorithm~\ref{alg:qdeq} monitors a stabilized relative residual for stopping. The unstructured forecasting map and the empirical $\sigma=1$ rule are evaluated numerically; their contractivity is not assumed. Positive-concave maps may also have guarantees under their own assumptions; our controlled positive-concave example additionally satisfies the contraction condition above.

\subsection{Acceptance and return policies}
\label{app:policies}
The safeguard uses the absolute residual norm $\|r\|$, while the stopping rule uses
\[\rho(z)=\|r(z)\|/(\|F(z)\|+10^{-9}).\]
 Table~\ref{tab:policies} defines the acceptance modes in Algorithm~\ref{alg:qdeq}. Damping changes the quadratic used to select a candidate and is separate from accepting that candidate. A rejected candidate triggers an explicitly evaluated Picard step $F(z)$.
\begin{table}[H]\centering\small
\caption{Safeguard modes. The strict test in the implementation implies the non-strict inequality used in the convergence proposition.}
\label{tab:policies}
\begin{tabular}{lp{.40\textwidth}p{.32\textwidth}}\toprule
Mode & Discrete candidate $z_c$ & Picard fallback \\\midrule
$\sigma=0$ & Apply the finite discrete update & Disabled \\
$\sigma=1$ & Accept when $\|r(z_c)\|<\|r(z)\|$ & Used on rejection \\
$0<\sigma<1$ & Accept when $\|r(z_c)\|<\sigma\|r(z)\|$ & Used on rejection \\
\bottomrule\end{tabular}\end{table}

The three-seed SA results and reference direction/precision ablations use $\sigma=0$ and the solver's final state. The paired horizon-96 safeguard study compares $\sigma=0$ with $\sigma=1$; the horizon-336 study additionally evaluates $\sigma=0.9$. These safeguard controls retain the final-state return policy. In the fresh CIM study, Weather, ETTh1, ETTh2 and ECL retain their original settings. The validation-selected Traffic hardware profile and its matched SA control use $\sigma=1$ and return the accepted outer iterate with the lowest measured relative residual, including the terminal iterate. The original Traffic SA reference is retained separately in Table~\ref{tab:cim}. Hard-budget controls specify their evaluated-state return policies in \cref{app:budget10,app:continuation}.

For a contractive $F$, Picard fallback and a fixed $\sigma<1$ give the geometric envelope proved above. For the unstructured forecasting maps, the safeguard controls measure empirical behavior. In particular, decrease-only candidate acceptance does not establish geometric convergence, and Picard fallback may increase the residual when $F$ is noncontractive.

\subsection{Implementation parameters}
The archived hardware adapter records its coupling scaling and spin dimensions with each submitted instance. The algorithm uses $b=10$ coefficient codes and 80 binary variables. Weather's continuous range estimate uses $a_{\rm cont}=-(G+\mu I)^{-1}h$, $\mu=\max(10^{-4}\operatorname{tr}(G)/m,2\times10^{-10})$, with a least-squares fallback. Fixed solver parameters and the range-controller margin were selected using validation performance. Dataset-specific settings are held fixed in the three-seed main table.

\section{Additional Solver Analyses}
\label{app:round3}

\subsection{Seeded validation-only selection} For each frozen seed-42 model, we cache eight validation windows using sampling seed 20261112 and evaluate the predeclared settings $(R/R_0,\eta/\eta_0)\in\{(1,1),(0.5,1),(2,1),(1,0.3),(1,3)\}$. The range multiplier acts on the dynamic margin for Weather, and the step multiplier acts on the secant ratio for ETTh2. We choose minimum validation MSE before constructing the 32-window test cohort. These inference-time sensitivities use frozen checkpoints and are reported separately from the three-seed main results.
\begin{table}[H]
\centering\small
\caption{Validation-selected solver sensitivities; test MSE on 32 held-out windows.}
\begin{adjustbox}{max width=\textwidth}
\begin{tabular}{lrrrrr}
\toprule
Dataset & $R/R_0$ & $\eta/\eta_0$ & Default & Selected & $\Delta$MSE (\%) \\
\midrule
weather & 1 & 0.3 & 0.181413 & 0.181421 & +0.004 \\
ETTh1 & 1 & 3 & 0.401474 & 0.401499 & +0.006 \\
ETTh2 & 1 & 0.3 & 0.308543 & 0.308439 & -0.034 \\
ECL & 1 & 0.3 & 0.144243 & 0.144445 & +0.141 \\
traffic & 0.5 & 1 & 0.392797 & 0.392649 & -0.038 \\
\bottomrule
\end{tabular}
\end{adjustbox}
\end{table}

\subsection{Synchronized runtime decomposition}
We synchronize the GPU around each timed bucket on 32 windows per dataset, separating probes, other network calls, QUBO backend, candidate decoding/scoring, basis and matrix construction. Times are instrumented host wall times; the sum of named buckets excludes small unassigned orchestration costs.
\begin{table}[H]
\centering\small
\caption{Instrumented batch-one SA inference components (ms/window).}
\begin{adjustbox}{max width=\textwidth}
\begin{tabular}{lrrrrrr}
\toprule
Dataset & Probes & Other network & SA backend & Candidate score & Basis+matrix & Total \\
\midrule
weather & 61.6 & 25.0 & 433.7 & 149.5 & 15.6 & 701.4 \\
ETTh1 & 50.1 & 18.9 & 367.0 & 126.2 & 11.2 & 587.6 \\
ETTh2 & 64.3 & 22.0 & 471.5 & 162.7 & 13.4 & 746.9 \\
ECL & 122.5 & 40.7 & 876.0 & 300.8 & 26.2 & 1388.4 \\
traffic & 69.7 & 23.8 & 497.7 & 171.6 & 15.4 & 791.9 \\
\bottomrule
\end{tabular}
\end{adjustbox}
\end{table}

\subsection{Strict ten-evaluation budget}
\label{app:budget10}
We impose nine shared-layer queries inside the solve and reserve one for re-forward. At budget exhaustion, all solvers return the evaluated state with the smallest measured residual. Perturbed probe states count as queries. This returns an evaluated state rather than applying an unchecked step; each of 32 windows is verified to use at most ten calls. Models are frozen and receive no additional distillation training.
\begin{table}[H]
\centering\small
\caption{MSE under a hard total budget of ten shared-layer forward evaluations.}
\begin{adjustbox}{max width=\textwidth}
\begin{tabular}{lrrrr}
\toprule
Dataset & Anderson & Broyden & Q-DEQ $m=8$ & Q-DEQ $m=4$ \\
\midrule
weather & 0.178088 & 0.184852 & 0.241006 & 0.219096 \\
ETTh1 & 0.397555 & 0.404695 & 0.410742 & 0.409840 \\
ETTh2 & 0.307551 & 0.308054 & 0.311955 & 0.311955 \\
ECL & 0.148459 & 0.150163 & 0.227008 & 0.204155 \\
traffic & 0.417169 & 0.406612 & 0.657515 & 0.659289 \\
\bottomrule
\end{tabular}
\end{adjustbox}
\end{table}

\subsection{Adaptive directions and missing-history stress}
On 16 paired windows per dataset, a greedy subset retains the residual direction and up to three additional original directions. Each new direction maximizes novelty after projection against the selected span; selection stops below novelty 0.1 once two directions are retained. A second stress test supplies no history, filling the remaining slots with the original random-direction rule. This is a controlled history-availability test, not a claim about an external out-of-distribution benchmark.
\begin{table}[H]
\centering\small
\caption{Direction ablation: task error, forward queries and history-stress convergence.}
\begin{adjustbox}{max width=\textwidth}
\begin{tabular}{lrrrrrr}
\toprule
Dataset & Reference MSE & Adaptive MSE & No-history MSE & Ref NFE & Adaptive NFE & No-history conv. \\
\midrule
weather & 0.167346 & 0.167398 & 0.167460 & 184.2 & 101.6 & 0.938 \\
ETTh1 & 0.376532 & 0.376560 & 0.376599 & 162.4 & 98.2 & 1.000 \\
ETTh2 & 0.202146 & 0.202038 & 0.202099 & 224.1 & 109.9 & 0.875 \\
ECL & 0.153580 & 0.153775 & 0.154285 & 374.4 & 232.1 & 0.000 \\
traffic & 0.374472 & 0.374458 & 0.374426 & 213.0 & 131.6 & 1.000 \\
\bottomrule
\end{tabular}
\end{adjustbox}
\end{table}

\subsection{Residual-adaptive bit allocation}
On 16 paired Weather and Traffic windows, we use $b=10$ while $\|r_k\|/\|r_0\|>0.3$, $b=6$ for ratios between 0.03 and 0.3, and $b=4$ thereafter. The same SA implementation and frozen checkpoint are used, with no hardware-service calls. All these runs converge at the default tolerance.
\begin{table}[H]
\centering\small
\caption{Residual-adaptive coefficient bits on 16 windows; wall time is a single-pass pilot.}
\begin{adjustbox}{max width=\textwidth}
\begin{tabular}{lrrrrr}
\toprule
Dataset & Fixed MSE & Adaptive MSE & Mean bits & Fixed s & Adaptive s \\
\midrule
weather & 0.188466 & 0.188442 & 5.886 & 0.699 & 0.636 \\
traffic & 0.381197 & 0.381138 & 6.215 & 0.771 & 0.686 \\
\bottomrule
\end{tabular}
\end{adjustbox}
\end{table}

\subsection{Training provenance and gradient snapshots}
The saved training loop logs test loss for monitoring but calls early stopping with validation loss. Across 33 complete training logs the loss records are finite; 22 runs stop before the ten-epoch cap and 11 reach it. These counts document the recorded training protocol. Solver comparisons use the selected checkpoints; the counts alone do not establish invariance to a longer training schedule. The table summarizes available logs; imported ETT headers do not contain complete Anderson/Q-DEQ learning curves. New snapshots use four training windows, a training-mode forward/backward and no optimizer update, comparing the same checkpoint with and without safeguards. They measure local gradient behavior and cost, not the outcome of a retrained safeguarded model.
\begin{table}[H]
\centering\small
\caption{Archived learning-curve summaries. Epoch times are historical log measurements, not a newly synchronized training benchmark.}
\begin{adjustbox}{max width=\textwidth}
\begin{tabular}{lrrrrr}
\toprule
Dataset & Solver & Runs & Epochs median & Best epoch median & s/epoch median \\
\midrule
weather & anderson & 3 & 6 & 3 & 86.8 \\
weather & broyden & 3 & 6 & 3 & 78.8 \\
weather & ising\_int8 & 3 & 6 & 3 & 1326.6 \\
ETTh1 & broyden & 3 & 5 & 2 & 12.8 \\
ETTh2 & broyden & 3 & 9 & 6 & 15.4 \\
ECL & anderson & 3 & 8 & 5 & 160.1 \\
ECL & broyden & 3 & 10 & 7 & 173.8 \\
ECL & ising\_int8 & 3 & 9 & 6 & 2338.2 \\
traffic & anderson & 3 & 10 & 10 & 528.9 \\
traffic & broyden & 3 & 10 & 10 & 440.7 \\
traffic & ising\_int8 & 3 & 10 & 10 & 2554.1 \\
\bottomrule
\end{tabular}
\end{adjustbox}
\end{table}\begin{table}[H]
\centering\small
\caption{Training-mode snapshot comparison over four fixed windows.}
\begin{adjustbox}{max width=\textwidth}
\begin{tabular}{lrrrrrrr}
\toprule
Dataset & Fallback off & Fallback on & NFE off & NFE on & s off & s on & Finite gradients \\
\midrule
weather & 0 & 4 & 185.5 & 181.5 & 0.737 & 0.707 & yes \\
ETTh1 & 0 & 0 & 160.5 & 160.5 & 0.618 & 0.616 & yes \\
ETTh2 & 0 & 0 & 138.0 & 138.0 & 0.539 & 0.537 & yes \\
ECL & 0 & 0 & 345.5 & 345.5 & 1.321 & 1.311 & yes \\
traffic & 0 & 0 & 183.0 & 183.0 & 0.719 & 0.712 & yes \\
\bottomrule
\end{tabular}
\end{adjustbox}
\end{table}

\subsection{Longer-horizon safeguards}
We load one frozen horizon-336 probe checkpoint per dataset and apply the corresponding horizon-96 seed-42 solver settings. Sixteen evenly spaced windows compare the default, decrease-only and uniform 0.9 residual test. This matched sensitivity experiment is separate from the main multi-horizon accuracy table.
\begin{table}[H]
\centering\small
\caption{Horizon-336 safeguard sensitivity (16 paired windows).}
\begin{adjustbox}{max width=\textwidth}
\begin{tabular}{lrrrrrr}
\toprule
Dataset & Policy & MSE & Converged & Mean residual & Mean steps & Fallbacks \\
\midrule
weather & off & 0.299563 & 0.438 & 0.02831 & 33.88 & 0 \\
weather & decrease & 0.300153 & 0.500 & 0.01044 & 34.25 & 232 \\
weather & uniform0.9 & 0.300437 & 0.562 & 0.00650 & 31.56 & 253 \\
ETTh1 & off & 0.476696 & 0.875 & 0.00208 & 19.19 & 0 \\
ETTh1 & decrease & 0.476896 & 0.875 & 0.00164 & 19.75 & 34 \\
ETTh1 & uniform0.9 & 0.476694 & 1.000 & 0.00073 & 17.50 & 45 \\
ETTh2 & off & 0.495066 & 1.000 & 0.00080 & 21.12 & 0 \\
ETTh2 & decrease & 0.495056 & 1.000 & 0.00083 & 20.88 & 12 \\
ETTh2 & uniform0.9 & 0.495073 & 1.000 & 0.00081 & 19.94 & 37 \\
ECL & off & 0.211542 & 0.875 & 0.00117 & 20.88 & 0 \\
ECL & decrease & 0.211541 & 0.875 & 0.00117 & 20.88 & 38 \\
ECL & uniform0.9 & 0.211544 & 0.875 & 0.00117 & 20.44 & 81 \\
traffic & off & 0.545511 & 1.000 & 0.00069 & 10.06 & 0 \\
traffic & decrease & 0.545486 & 1.000 & 0.00070 & 10.31 & 5 \\
traffic & uniform0.9 & 0.545524 & 1.000 & 0.00068 & 10.00 & 4 \\
\bottomrule
\end{tabular}
\end{adjustbox}
\end{table}

\subsection{Partial quantization inside the solve}
On 16 paired ETTh2 and Traffic windows, we compare the original re-forward-only design, dynamic per-tensor activation quantization inside the solve, per-channel activation scaling, and weight-only quantization with floating-point activations. Linear activations use the feature axis; Conv1d activations use the channel axis. All variants use symmetric 8-bit weights and the same frozen checkpoint and sample order. The partial-quantization module replacements remain active for both the solve and terminal call: the weight-only regime uses floating-point activations at both stages, while the reference uses FP32 solving and a W8A8 terminal re-forward.
\begin{table}[H]
\centering\small
\caption{Partial shared-layer quantization sensitivities on 16 paired windows per dataset. W8A32 denotes quantized weights and floating-point activations.}
\label{tab:partial-quant}
\begin{adjustbox}{max width=\textwidth}
\begin{tabular}{lrrrr}
\toprule
Dataset & Mode & MSE & Converged & Mean steps \\
\midrule
ETTh2 & FP32 solve / W8A8 terminal & 0.202216 & 0.938 & 20.00 \\
ETTh2 & W8A8, per-tensor activations & 0.201898 & 0.000 & 41.00 \\
ETTh2 & W8A8, per-channel activations & 0.202131 & 0.750 & 28.56 \\
ETTh2 & W8A32 & 0.202100 & 0.875 & 21.62 \\
traffic & FP32 solve / W8A8 terminal & 0.381197 & 1.000 & 22.31 \\
traffic & W8A8, per-tensor activations & 0.589954 & 0.000 & 41.00 \\
traffic & W8A8, per-channel activations & 0.466093 & 0.000 & 41.00 \\
traffic & W8A32 & 0.380661 & 1.000 & 22.44 \\
\bottomrule
\end{tabular}
\end{adjustbox}
\end{table}

\subsection{Controlled structured-map experiment}
We test two 16-dimensional map families over 20 seeds each: $F(z)=\tanh(Wz+u)$ with $\|W\|_2=0.6$, and the positive-concave map $F(z)=\sqrt{Wz+\mathbf1}$ for $W\geq0$, $\|W\|_2=0.6$ and $z\geq0$. Their contraction bounds are 0.6 and 0.3. A two-direction, four-bit-per-direction QUBO is solved by exhaustive enumeration, with domain checking and the $\gamma=0.9$ safeguard. This isolates discrete-step compatibility and does not substitute for training the published monDEQ/pcDEQ architectures.
\begin{table}[H]
\centering\small
\caption{Structured-map checks: all 80 runs reach absolute residual below $10^{-6}$; every safeguarded trajectory satisfies its geometric envelope.}
\begin{adjustbox}{max width=\textwidth}
\begin{tabular}{lrrrr}
\toprule
Map & Solver & Mean iterations & Mean NFE & Max residual \\
\midrule
contractive\_tanh & picard & 12.35 & 13.35 & 9.68e-07 \\
contractive\_tanh & discrete\_safeguarded & 11.60 & 47.40 & 9.36e-07 \\
positive\_concave\_sqrt & picard & 10.00 & 11.00 & 3.12e-07 \\
positive\_concave\_sqrt & discrete\_safeguarded & 7.50 & 31.00 & 9.89e-07 \\
\bottomrule
\end{tabular}
\end{adjustbox}
\end{table}

\section{Measured CIM Task Records}
\label{app:measured-cim}
The paired rerun uses the original fixed checkpoints and first 128 test windows per dataset. Each saved successful task has a unique task identifier, input-matrix hash, ten returned configurations and its SDK \texttt{time\_consume} value. The task timing is read directly in microseconds; repeated polling records are deduplicated by task identifier. The aggregate per-window time equals the sum of completed task times divided by 128 and is amortized at the recorded batch size. The pilot task is excluded from these statistics. Forecast targets agree elementwise between the SA and CIM cohorts, and reported errors are recomputed from saved forecasts.
\begin{table}[H]
\centering\small
\caption{Original SA reference MAE, CIM hardware MAE, and measured task-time statistics in microseconds. Dispersion is across the unique successful tasks, not across training seeds.}
\begin{adjustbox}{max width=\textwidth}
\begin{tabular}{lrrrrrr}
\toprule
Dataset & SA MAE & CIM MAE & Tasks & Mean & Std & Median \\
\midrule
Weather & 0.3044 & 0.3043 & 111 & 3318.9 & 2928.5 & 2107.8 \\
ETTh1 & 0.3511 & 0.3511 & 82 & 3787.8 & 3093.2 & 2530.0 \\
ETTh2 & 0.2095 & 0.2090 & 150 & 4582.1 & 3130.8 & 3723.0 \\
ECL & 0.2101 & 0.2101 & 320 & 4713.1 & 4615.2 & 2779.3 \\
Traffic & 0.2350 & 0.2346 & 273 & 4577.8 & 3665.0 & 3660.6 \\
\bottomrule
\end{tabular}
\end{adjustbox}
\end{table}

\subsection{Traffic hardware profile}
\label{app:traffic-profile}
The coefficient grid remains at eight directions and ten bits per direction. The submitted matrix is symmetrized with zero diagonal, scaled by its maximum absolute entry, multiplied by 8191, and rounded to an integer. The sign is negated for the SDK's maximization convention. This keeps the 80 coefficient spins and one reference spin. Returned samples are decoded and ranked in the original local quadratic. No software annealing or local-search refinement is applied to the physical returns.

The profile is selected by minimum CIM forecast MSE on 16 validation windows sampled without replacement using seed 20261117. Candidate repairs vary coefficient resolution (four, six or ten bits), residual-decrease protection, the returned iterate, and the coupling representation. Promotion additionally requires the validation MSE to stay within one percent of both the paired SA reference and the original ten-bit SA reference. The selected profile is then evaluated once on all of the original first 128 test windows. The eight batches of 16 are evaluated independently with the original per-batch direction randomness and solver seed offsets; a replay of all 320 original tasks verifies identical input hashes and seeds under this execution arrangement. Device timing sums the unique successful task records across all eight batches.

The SA control under the selected host profile gives MSE 0.4201814 and MAE 0.2458463; CIM gives MSE 0.4090492 and MAE 0.2346285. The original SA reference retained in \cref{tab:cim} gives MSE 0.4094488 and MAE 0.2349649. The main forecasting baselines and training are unchanged.

\section{Matched Budgets and Learned Acceleration}
\label{app:continuation}
\begin{table}[H]\centering\small
\caption{Computation-limited forecasts on 256 windows per dataset, teacher seed 42. Entries are MSE / actual backbone queries under a total ceiling of 30 calls. All methods use the same frozen Q-DEQ checkpoint or its distilled student. C-DEQ depth is selected by validation MSE among the prelisted depths fitting the ceiling. Student training and the time readout add parameters and computation; query counts are not equal-FLOP budgets.}
\label{tab:budget-main}
\begin{adjustbox}{max width=\textwidth}
\begin{tabular}{lccccc}\toprule
Dataset & Broyden & Q-DEQ $m=8$ & Adaptive Q-DEQ & C-DEQ adaptation & Warm start + Q-DEQ \\\midrule
Weather & 0.1756 / 18.4 & 0.1971 / 30.0 & 0.1763 / 30.0 & 0.1765 / 5.0 & 0.1752 / 30.0 \\
ETTh1 & 0.3978 / 21.5 & 0.4062 / 30.0 & 0.3994 / 30.0 & 0.3975 / 3.0 & 0.3982 / 30.0 \\
ETTh2 & 0.2962 / 20.2 & 0.2998 / 30.0 & 0.2966 / 30.0 & 0.2972 / 5.0 & 0.2966 / 30.0 \\
ECL & 0.1489 / 29.9 & 0.1935 / 30.0 & 0.1684 / 30.0 & 0.1558 / 5.0 & 0.1523 / 30.0 \\
Traffic & 0.3981 / 20.1 & 0.6748 / 30.0 & 0.5775 / 30.0 & 0.4148 / 17.0 & 0.4156 / 30.0 \\
\bottomrule\end{tabular}\end{adjustbox}\end{table}

\subsection{Protocol and actual query accounting}
The continuation fixes the seed-42 checkpoints and 256 evenly spaced individual test windows per dataset. All eight predeclared solver/checkpoint configurations are evaluated at total caps 25 and 30. The network-call hook independently verifies the query meter on every window. Finite-difference perturbations count as queries, and one call is reserved for the terminal W8A8 re-forward of the Q-DEQ checkpoint. The selected returned state minimizes the measured relative residual among all queried states, including perturbation probes, under both normal termination and budget exhaustion. This is a common budgeted return policy, separate from the historical final-state policy of the main SA table.

Native Anderson and Broyden retain their own trained checkpoints and FP32 output conventions. The shared-map solver swaps use the Q-DEQ weights, common input injection and W8A8 terminal representation. All Q-DEQ step controls retain their recorded dataset settings. The adaptive rule retains the residual direction and greedily selects at most three additional directions by two-pass Gram--Schmidt novelty, stopping below 0.1 after retaining two directions unless rank is exhausted. No candidate configuration is selected using test MSE.

Each setting uses one evaluation pass for error and query-count measurement. Synchronized wall times from these shared-GPU runs are retained as execution diagnostics; isolated runtime measurements and physical device task records use their separately stated protocols.
\begingroup\fontsize{8.5}{10}\selectfont\setlength{\tabcolsep}{3.5pt}
\setlength{\LTpre}{7pt}\setlength{\LTpost}{7pt}
\begin{longtable}{llllll}
\caption{Total query ceiling 25: forecasts, actual mean backbone calls and fraction reaching relative residual below $10^{-3}$. Native and shared-map checkpoints are identified separately. The near-cap C-DEQ setting uses 12 consistency steps and 25 actual calls; validation-selected depth may use fewer.}\label{tab:budget-25}\\
\toprule
Dataset & Method & MSE & MAE & Queries & Converged \\\midrule\endfirsthead
\toprule
Dataset & Method & MSE & MAE & Queries & Converged \\\midrule\endhead
\midrule\multicolumn{6}{r}{\footnotesize Continued on next page}\\\endfoot
\bottomrule\endlastfoot

Weather & Native AA & 0.177323 & 0.215775 & 18.74 & 0.879 \\
Weather & Native Broyden & 0.178153 & 0.215390 & 21.35 & 0.738 \\
Weather & Shared AA & 0.175588 & 0.215448 & 17.80 & 0.941 \\
Weather & Shared Broyden & 0.175627 & 0.215499 & 18.23 & 0.934 \\
Weather & Q-DEQ m=2 & 0.174240 & 0.214884 & 25.00 & 0.000 \\
Weather & Q-DEQ m=4 & 0.176165 & 0.216601 & 25.00 & 0.000 \\
Weather & Q-DEQ m=8 & 0.197097 & 0.235252 & 25.00 & 0.000 \\
Weather & Adaptive Q-DEQ & 0.176299 & 0.216401 & 25.00 & 0.000 \\
Weather & C-DEQ (near cap) & 0.180123 & 0.219220 & 25.00 & 0.000 \\
Weather & C-DEQ (val. depth) & 0.176491 & 0.216304 & 5.00 & 0.000 \\
Weather & Learned warm start + Q-DEQ & 0.175231 & 0.215147 & 25.00 & 0.000 \\
ETTh1 & Native AA & 0.396032 & 0.409235 & 20.75 & 0.773 \\
ETTh1 & Native Broyden & 0.399540 & 0.411769 & 22.27 & 0.660 \\
ETTh1 & Shared AA & 0.397836 & 0.410387 & 18.82 & 0.969 \\
ETTh1 & Shared Broyden & 0.397781 & 0.410378 & 21.18 & 0.871 \\
ETTh1 & Q-DEQ m=2 & 0.397811 & 0.410385 & 25.00 & 0.000 \\
ETTh1 & Q-DEQ m=4 & 0.399092 & 0.411607 & 25.00 & 0.000 \\
ETTh1 & Q-DEQ m=8 & 0.406232 & 0.415112 & 25.00 & 0.000 \\
ETTh1 & Adaptive Q-DEQ & 0.399598 & 0.411833 & 25.00 & 0.000 \\
ETTh1 & C-DEQ (near cap) & 0.398469 & 0.410450 & 25.00 & 0.000 \\
ETTh1 & C-DEQ (val. depth) & 0.397511 & 0.410063 & 3.00 & 0.000 \\
ETTh1 & Learned warm start + Q-DEQ & 0.398262 & 0.410748 & 25.00 & 0.000 \\
ETTh2 & Native AA & 0.296072 & 0.349023 & 18.66 & 0.766 \\
ETTh2 & Native Broyden & 0.296807 & 0.349711 & 19.56 & 0.816 \\
ETTh2 & Shared AA & 0.296088 & 0.349621 & 17.92 & 0.828 \\
ETTh2 & Shared Broyden & 0.296156 & 0.349708 & 19.43 & 0.805 \\
ETTh2 & Q-DEQ m=2 & 0.296385 & 0.350139 & 25.00 & 0.000 \\
ETTh2 & Q-DEQ m=4 & 0.296371 & 0.350269 & 25.00 & 0.000 \\
ETTh2 & Q-DEQ m=8 & 0.299786 & 0.351778 & 25.00 & 0.000 \\
ETTh2 & Adaptive Q-DEQ & 0.296559 & 0.350337 & 25.00 & 0.000 \\
ETTh2 & C-DEQ (near cap) & 0.299623 & 0.351184 & 25.00 & 0.000 \\
ETTh2 & C-DEQ (val. depth) & 0.297210 & 0.350091 & 5.00 & 0.000 \\
ETTh2 & Learned warm start + Q-DEQ & 0.296566 & 0.349892 & 25.00 & 0.000 \\
ECL & Native AA & 0.151411 & 0.243295 & 25.00 & 0.000 \\
ECL & Native Broyden & 0.151553 & 0.242595 & 25.00 & 0.000 \\
ECL & Shared AA & 0.148935 & 0.239914 & 25.00 & 0.000 \\
ECL & Shared Broyden & 0.148841 & 0.239891 & 25.00 & 0.000 \\
ECL & Q-DEQ m=2 & 0.158878 & 0.249643 & 25.00 & 0.000 \\
ECL & Q-DEQ m=4 & 0.168005 & 0.259410 & 25.00 & 0.000 \\
ECL & Q-DEQ m=8 & 0.193479 & 0.285601 & 25.00 & 0.000 \\
ECL & Adaptive Q-DEQ & 0.168427 & 0.259697 & 25.00 & 0.000 \\
ECL & C-DEQ (near cap) & 0.163007 & 0.253241 & 25.00 & 0.000 \\
ECL & C-DEQ (val. depth) & 0.155750 & 0.245063 & 5.00 & 0.000 \\
ECL & Learned warm start + Q-DEQ & 0.152330 & 0.241694 & 25.00 & 0.000 \\
Traffic & Native AA & 0.400241 & 0.271217 & 24.09 & 0.500 \\
Traffic & Native Broyden & 0.395754 & 0.268949 & 22.05 & 0.879 \\
Traffic & Shared AA & 0.398254 & 0.270180 & 21.39 & 0.887 \\
Traffic & Shared Broyden & 0.398123 & 0.270075 & 19.97 & 0.965 \\
Traffic & Q-DEQ m=2 & 0.520401 & 0.373743 & 25.00 & 0.000 \\
Traffic & Q-DEQ m=4 & 0.582325 & 0.424763 & 25.00 & 0.000 \\
Traffic & Q-DEQ m=8 & 0.674772 & 0.482791 & 25.00 & 0.000 \\
Traffic & Adaptive Q-DEQ & 0.577559 & 0.419650 & 25.00 & 0.000 \\
Traffic & C-DEQ (near cap) & 0.414481 & 0.285646 & 25.00 & 0.000 \\
Traffic & C-DEQ (val. depth) & 0.414803 & 0.285657 & 17.00 & 0.000 \\
Traffic & Learned warm start + Q-DEQ & 0.415608 & 0.283230 & 25.00 & 0.000 \\
\end{longtable}\endgroup

\begingroup\fontsize{8.5}{10}\selectfont\setlength{\tabcolsep}{3.5pt}
\setlength{\LTpre}{7pt}\setlength{\LTpost}{7pt}
\begin{longtable}{llllll}
\caption{Total query ceiling 30: forecasts, actual mean backbone calls and fraction reaching relative residual below $10^{-3}$. Native and shared-map checkpoints are identified separately. The near-cap C-DEQ setting uses 14 consistency steps and 29 actual calls; validation-selected depth may use fewer.}\label{tab:budget-30}\\
\toprule
Dataset & Method & MSE & MAE & Queries & Converged \\\midrule\endfirsthead
\toprule
Dataset & Method & MSE & MAE & Queries & Converged \\\midrule\endhead
\midrule\multicolumn{6}{r}{\footnotesize Continued on next page}\\\endfoot
\bottomrule\endlastfoot

Weather & Native AA & 0.177355 & 0.215802 & 19.22 & 0.926 \\
Weather & Native Broyden & 0.178242 & 0.215560 & 22.36 & 0.855 \\
Weather & Shared AA & 0.175605 & 0.215471 & 18.02 & 0.977 \\
Weather & Shared Broyden & 0.175636 & 0.215508 & 18.41 & 0.988 \\
Weather & Q-DEQ m=2 & 0.174113 & 0.214687 & 30.00 & 0.000 \\
Weather & Q-DEQ m=4 & 0.176124 & 0.216580 & 30.00 & 0.000 \\
Weather & Q-DEQ m=8 & 0.197097 & 0.235253 & 30.00 & 0.000 \\
Weather & Adaptive Q-DEQ & 0.176272 & 0.216372 & 30.00 & 0.000 \\
Weather & C-DEQ (near cap) & 0.180163 & 0.219273 & 29.00 & 0.000 \\
Weather & C-DEQ (val. depth) & 0.176491 & 0.216304 & 5.00 & 0.000 \\
Weather & Learned warm start + Q-DEQ & 0.175216 & 0.215135 & 30.00 & 0.000 \\
ETTh1 & Native AA & 0.396028 & 0.409244 & 21.63 & 0.875 \\
ETTh1 & Native Broyden & 0.399528 & 0.411755 & 23.50 & 0.848 \\
ETTh1 & Shared AA & 0.397837 & 0.410388 & 18.90 & 0.992 \\
ETTh1 & Shared Broyden & 0.397772 & 0.410376 & 21.52 & 0.980 \\
ETTh1 & Q-DEQ m=2 & 0.397711 & 0.410184 & 30.00 & 0.000 \\
ETTh1 & Q-DEQ m=4 & 0.399088 & 0.411608 & 30.00 & 0.000 \\
ETTh1 & Q-DEQ m=8 & 0.406218 & 0.415101 & 30.00 & 0.000 \\
ETTh1 & Adaptive Q-DEQ & 0.399408 & 0.411548 & 30.00 & 0.000 \\
ETTh1 & C-DEQ (near cap) & 0.398559 & 0.410506 & 29.00 & 0.000 \\
ETTh1 & C-DEQ (val. depth) & 0.397511 & 0.410063 & 3.00 & 0.000 \\
ETTh1 & Learned warm start + Q-DEQ & 0.398173 & 0.410681 & 30.00 & 0.000 \\
ETTh2 & Native AA & 0.296240 & 0.349140 & 19.70 & 0.816 \\
ETTh2 & Native Broyden & 0.296870 & 0.349721 & 20.30 & 0.887 \\
ETTh2 & Shared AA & 0.296132 & 0.349646 & 18.72 & 0.855 \\
ETTh2 & Shared Broyden & 0.296174 & 0.349739 & 20.23 & 0.883 \\
ETTh2 & Q-DEQ m=2 & 0.296461 & 0.350053 & 30.00 & 0.000 \\
ETTh2 & Q-DEQ m=4 & 0.296371 & 0.350269 & 30.00 & 0.000 \\
ETTh2 & Q-DEQ m=8 & 0.299786 & 0.351778 & 30.00 & 0.000 \\
ETTh2 & Adaptive Q-DEQ & 0.296559 & 0.350337 & 30.00 & 0.000 \\
ETTh2 & C-DEQ (near cap) & 0.299671 & 0.351160 & 29.00 & 0.000 \\
ETTh2 & C-DEQ (val. depth) & 0.297210 & 0.350091 & 5.00 & 0.000 \\
ETTh2 & Learned warm start + Q-DEQ & 0.296563 & 0.349890 & 30.00 & 0.000 \\
ECL & Native AA & 0.151508 & 0.243385 & 30.00 & 0.000 \\
ECL & Native Broyden & 0.151926 & 0.242888 & 29.96 & 0.043 \\
ECL & Shared AA & 0.148953 & 0.239933 & 29.98 & 0.035 \\
ECL & Shared Broyden & 0.148878 & 0.239901 & 29.87 & 0.090 \\
ECL & Q-DEQ m=2 & 0.156271 & 0.247160 & 30.00 & 0.000 \\
ECL & Q-DEQ m=4 & 0.167980 & 0.259385 & 30.00 & 0.000 \\
ECL & Q-DEQ m=8 & 0.193477 & 0.285600 & 30.00 & 0.000 \\
ECL & Adaptive Q-DEQ & 0.168397 & 0.259650 & 30.00 & 0.000 \\
ECL & C-DEQ (near cap) & 0.163827 & 0.253870 & 29.00 & 0.000 \\
ECL & C-DEQ (val. depth) & 0.155750 & 0.245063 & 5.00 & 0.000 \\
ECL & Learned warm start + Q-DEQ & 0.152340 & 0.241680 & 30.00 & 0.000 \\
Traffic & Native AA & 0.400449 & 0.271386 & 25.65 & 0.875 \\
Traffic & Native Broyden & 0.395768 & 0.268958 & 22.53 & 0.938 \\
Traffic & Shared AA & 0.398334 & 0.270256 & 21.73 & 0.973 \\
Traffic & Shared Broyden & 0.398137 & 0.270079 & 20.08 & 0.988 \\
Traffic & Q-DEQ m=2 & 0.490185 & 0.350322 & 30.00 & 0.000 \\
Traffic & Q-DEQ m=4 & 0.582230 & 0.424683 & 30.00 & 0.000 \\
Traffic & Q-DEQ m=8 & 0.674811 & 0.482852 & 30.00 & 0.000 \\
Traffic & Adaptive Q-DEQ & 0.577536 & 0.419621 & 30.00 & 0.000 \\
Traffic & C-DEQ (near cap) & 0.414262 & 0.285500 & 29.00 & 0.000 \\
Traffic & C-DEQ (val. depth) & 0.414803 & 0.285657 & 17.00 & 0.000 \\
Traffic & Learned warm start + Q-DEQ & 0.415578 & 0.283208 & 30.00 & 0.000 \\
\end{longtable}\endgroup

\begin{figure}[!htbp]\centering
\includegraphics[width=\textwidth,height=0.68\textheight,keepaspectratio]{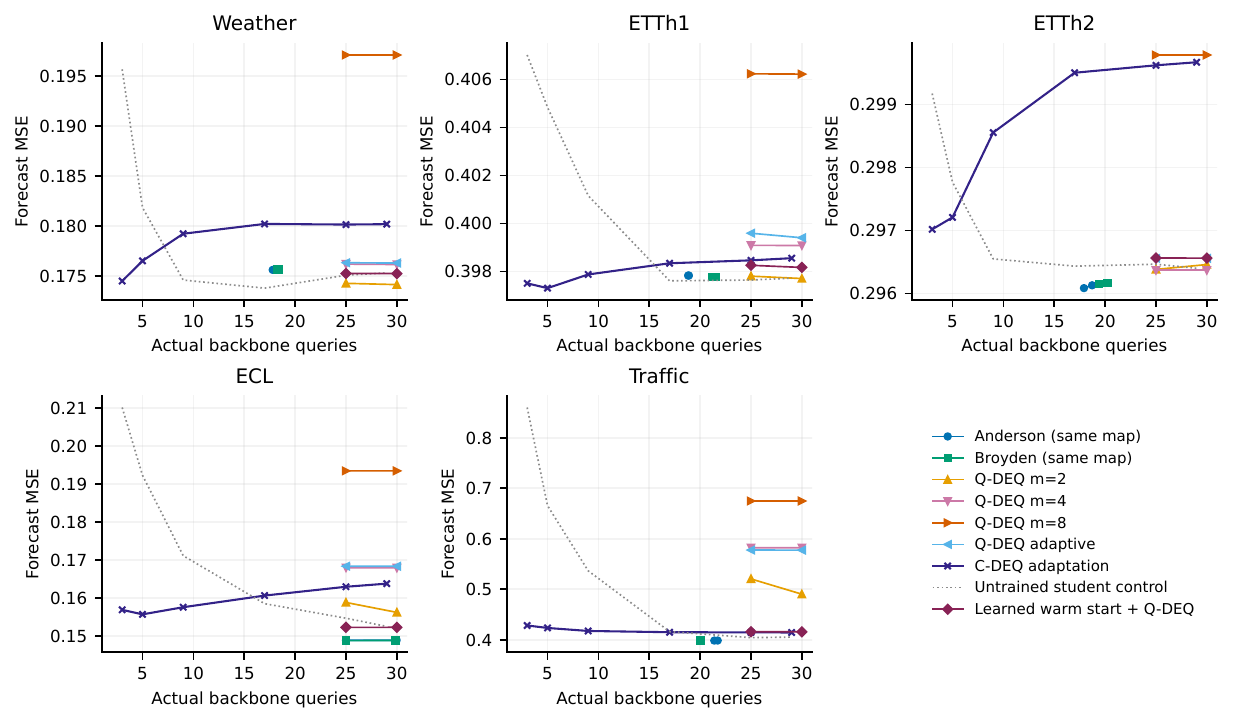}
\caption{Forecast error and actual backbone queries on the same frozen Q-DEQ maps. All predeclared direction settings and consistency-step counts are shown, including the untrained student-copy control. Native-checkpoint baselines appear separately in the tables. Student readout work and additional training are reported independently of the backbone query count.}
\label{fig:matched-budget}\end{figure}

\subsection{Larger-cohort adaptive direction study}
The reference and adaptive solvers use the same checkpoint, inputs, per-window seeds, acceptance settings and lowest-observed-residual return rule. Their outer cap is 40 and the query ceiling is 1,024, which is nonbinding. The experiment measures the effect of selecting a smaller subspace under this common return rule.
\begin{table}[H]\centering\small
\caption{Adaptive directions on 256 paired windows per dataset. Forecast MSE, mean query count and tolerance attainment are reported together.}
\label{tab:adaptive-large}
\begin{adjustbox}{max width=\textwidth}
\begin{tabular}{lllllll}\toprule
Dataset & Ref. MSE & Adaptive MSE & Ref. queries & Adaptive queries & Ref. conv. & Adaptive conv. \\\midrule
Weather & 0.175631 & 0.175621 & 181.9 & 104.6 & 0.984 & 0.988 \\
ETTh1 & 0.397781 & 0.397825 & 156.0 & 93.8 & 1.000 & 0.996 \\
ETTh2 & 0.296530 & 0.296222 & 189.9 & 95.2 & 0.945 & 0.953 \\
ECL & 0.148970 & 0.148979 & 351.3 & 219.0 & 0.742 & 0.621 \\
Traffic & 0.397912 & 0.397890 & 210.7 & 131.0 & 0.988 & 0.980 \\
\bottomrule\end{tabular}\end{adjustbox}\end{table}

\subsection{C-DEQ forecasting adaptation and training}
The implementation follows the consistency-distillation formulation of \citet{lin2026consistency}, with its two-state Anderson operation checked numerically against the authors' public implementation at commit \texttt{bcd0e51cf8b3a1d9e0405a88ca2c581b00dd11fc}. The forecasting backbone and task readout are adapted to iTransformer; these are newly trained forecasting students, rather than the authors' released NLP, vision or graph checkpoints.

For embedded input $x$, coordinates $w=z-x$ give initial state $w_0=0$ and shifted map $F(w+x,x)-x$. The student has a separately trainable copy of the shared layer and injection normalization. A two-state AA mixture feeds a time-conditioned $(d+1)\to3d\to d$ ReLU readout, initialized to act as the identity. The boundary coefficients are $c_{\rm skip}(t)=t^2$ and $c_{\rm out}(t)=1-t^2$ in normalized time. Training times use $t_k=1-e^{-0.1k}$, and inference uses $t_j=1-0.9^j$.

We cache FP32 teacher trajectories using 80 Anderson queries per training input, choosing each input's lowest-measured-residual state as its endpoint. Each dataset uses 1,024 fixed training windows and 128 fixed validation windows from the original chronological splits. Thirty percent of cached training pairs use the initial state; remaining pairs sample intermediate adjacent states, with the stated endpoint augmentation probability 0.1 away from trajectory boundaries. Distillation minimizes 0.5 global endpoint MSE plus 0.5 adjacent-time EMA consistency MSE plus 0.01 forecast-task MSE through the frozen readout. Adam uses learning rate $3\times10^{-4}$, batch eight, gradient clipping one, EMA decay 0.99, a cap of 30 epochs and patience five. Dropout is disabled for the deterministic consistency maps. The selected epoch minimizes validation forecast MSE at four consistency steps. With that checkpoint fixed, inference depth minimizes batch-one validation MSE among $J\in\{1,2,4,8,12,14\}$ fitting the query ceiling. This allows fewer calls than the ceiling; test metrics do not select the depth. All near-cap and full-curve results remain reported.

Teacher trajectory quality and added student parameters appear in \cref{tab:student-training}. The original teacher layer is retained for the common terminal re-forward. Each inference uses one student backbone call at the first consistency step and two thereafter, followed by one teacher FP32 residual check and one W8A8 terminal re-forward: $2J+1$ calls for $J$ steps. Hooks verify these counts. The time readout adds operations beyond those backbone calls. Thus equal query ceilings describe function-evaluation budgets and do not imply equal FLOPs or training cost. The warm-start variant uses one student step and reserves all remaining calls for Q-DEQ probes, residuals and the terminal re-forward.
\begin{table}[H]\centering\small
\caption{Distillation protocol outcomes. Teacher convergence is the training-window fraction reaching $10^{-3}$ within 80 queries. Epochs are counted from one; parameter counts are in millions. Teacher and student parameters are separate.}
\label{tab:student-training}
\begin{adjustbox}{max width=\textwidth}
\begin{tabular}{llllll}\toprule
Dataset & Teacher conv. & Epochs & Selected epoch & Student params & Retained teacher params \\\midrule
Weather & 0.991 & 14 & 9 & 3.155 & 1.679 \\
ETTh1 & 0.991 & 16 & 11 & 0.791 & 0.446 \\
ETTh2 & 0.893 & 7 & 2 & 0.199 & 0.125 \\
ECL & 0.910 & 25 & 20 & 3.155 & 1.679 \\
Traffic & 1.000 & 9 & 4 & 3.155 & 1.679 \\
\bottomrule\end{tabular}\end{adjustbox}\end{table}
\begin{table}[H]\centering\small
\caption{Validation-selected inference depth. The selected depths fit both ceilings 25 and 30 for these checkpoints. The time-readout work is additional to the listed backbone calls.}
\label{tab:depth-selection}
\begin{adjustbox}{max width=\textwidth}
\begin{tabular}{lllll}\toprule
Dataset & Steps & Actual calls & Validation MSE & Test MSE \\\midrule
Weather & 2 & 5 & 0.391374 & 0.176491 \\
ETTh1 & 1 & 3 & 0.676780 & 0.397511 \\
ETTh2 & 2 & 5 & 0.220560 & 0.297210 \\
ECL & 2 & 5 & 0.136401 & 0.155750 \\
Traffic & 8 & 17 & 0.378758 & 0.414803 \\
\bottomrule\end{tabular}\end{adjustbox}\end{table}
\begingroup\fontsize{8.5}{10}\selectfont\setlength{\tabcolsep}{3.5pt}
\setlength{\LTpre}{7pt}\setlength{\LTpost}{7pt}
\begin{longtable}{lllllll}
\caption{All consistency-step settings, including the untrained teacher-copy control with the identity time readout. The complete curve is retained even when extra consistency steps increase forecast error.}\label{tab:consistency-curves}\\
\toprule
Dataset & Steps & Queries & Untrained MSE & Distilled MSE & Distilled MAE & Converged \\\midrule\endfirsthead
\toprule
Dataset & Steps & Queries & Untrained MSE & Distilled MSE & Distilled MAE & Converged \\\midrule\endhead
\midrule\multicolumn{7}{r}{\footnotesize Continued on next page}\\\endfoot
\bottomrule\endlastfoot

Weather & 1 & 3 & 0.195662 & 0.174466 & 0.214989 & 0.000 \\
Weather & 2 & 5 & 0.181819 & 0.176491 & 0.216304 & 0.000 \\
Weather & 4 & 9 & 0.174568 & 0.179205 & 0.218322 & 0.000 \\
Weather & 8 & 17 & 0.173759 & 0.180187 & 0.219144 & 0.000 \\
Weather & 12 & 25 & 0.175050 & 0.180123 & 0.219220 & 0.000 \\
Weather & 14 & 29 & 0.175282 & 0.180163 & 0.219273 & 0.000 \\
ETTh1 & 1 & 3 & 0.407015 & 0.397511 & 0.410063 & 0.000 \\
ETTh1 & 2 & 5 & 0.404839 & 0.397312 & 0.409986 & 0.000 \\
ETTh1 & 4 & 9 & 0.401160 & 0.397881 & 0.410151 & 0.000 \\
ETTh1 & 8 & 17 & 0.397615 & 0.398347 & 0.410394 & 0.000 \\
ETTh1 & 12 & 25 & 0.397650 & 0.398469 & 0.410450 & 0.000 \\
ETTh1 & 14 & 29 & 0.397727 & 0.398559 & 0.410506 & 0.000 \\
ETTh2 & 1 & 3 & 0.299174 & 0.297020 & 0.349846 & 0.000 \\
ETTh2 & 2 & 5 & 0.297768 & 0.297210 & 0.350091 & 0.000 \\
ETTh2 & 4 & 9 & 0.296551 & 0.298556 & 0.350738 & 0.000 \\
ETTh2 & 8 & 17 & 0.296437 & 0.299504 & 0.351052 & 0.000 \\
ETTh2 & 12 & 25 & 0.296467 & 0.299623 & 0.351184 & 0.000 \\
ETTh2 & 14 & 29 & 0.296403 & 0.299671 & 0.351160 & 0.000 \\
ECL & 1 & 3 & 0.210154 & 0.156941 & 0.244929 & 0.000 \\
ECL & 2 & 5 & 0.192328 & 0.155750 & 0.245063 & 0.000 \\
ECL & 4 & 9 & 0.171174 & 0.157625 & 0.247967 & 0.000 \\
ECL & 8 & 17 & 0.158562 & 0.160689 & 0.251266 & 0.000 \\
ECL & 12 & 25 & 0.154737 & 0.163007 & 0.253241 & 0.000 \\
ECL & 14 & 29 & 0.152583 & 0.163827 & 0.253870 & 0.000 \\
Traffic & 1 & 3 & 0.860121 & 0.428178 & 0.294034 & 0.000 \\
Traffic & 2 & 5 & 0.665703 & 0.423231 & 0.292085 & 0.000 \\
Traffic & 4 & 9 & 0.536464 & 0.417286 & 0.287049 & 0.000 \\
Traffic & 8 & 17 & 0.415967 & 0.414803 & 0.285657 & 0.000 \\
Traffic & 12 & 25 & 0.404030 & 0.414481 & 0.285646 & 0.000 \\
Traffic & 14 & 29 & 0.405074 & 0.414262 & 0.285500 & 0.000 \\
\end{longtable}\endgroup

\end{document}